\documentclass[runningheads]{llncs}

\usepackage{eccv}

\usepackage{eccvabbrv}
\usepackage{tcolorbox} 

\usepackage{CJKutf8}

\usepackage{graphicx}
\usepackage{booktabs}

\usepackage[accsupp]{axessibility}  

\usepackage[pagebackref,breaklinks,colorlinks,citecolor=eccvblue]{hyperref}

\usepackage{orcidlink}

\begin{document}

\title{MEDCO: Medical Education Copilots Based on A Multi-Agent Framework} 

\titlerunning{MEDCO: Medical Education Copilot}

\author{Hao Wei\inst{1}\orcidlink{0000-1111-2222-3333} \and
Jianing Qiu\inst{1}\orcidlink{0000-0003-4166-3428} \and
Haibao Yu\inst{2}\orcidlink{0000-1111-2222-3333} \and
Wu Yuan\inst{1}\orcidlink{0000-0001-9405-519X}}

\authorrunning{H.~Wei et al.}

\institute{$^1$ The Chinese University of Hong Kong, Hong Kong SAR\\
$^2$ The University of Hong Kong, Hong Kong SAR\\
\email{haowei@link.cuhk.edu.hk, yuhaibao@connect.hku.hk, \\\{jianingqiu,wyuan\}@cuhk.edu.hk}}

\maketitle

\begin{abstract}

Large language models (LLMs) have had a significant impact on diverse research domains, including medicine and healthcare. However, the potential of LLMs as copilots in medical education remains underexplored. Current AI-assisted educational tools are limited by their solitary learning approach and inability to simulate the multi-disciplinary and interactive nature of actual medical training. To address these limitations, we propose MEDCO (Medical EDucation COpilots), a novel multi-agent-based copilot system specially developed to emulate real-world medical training environments. MEDCO incorporates three primary agents: an agentic patient, an expert doctor, and a radiologist, facilitating a multi-modal and interactive learning environment. Our framework emphasizes the learning of proficient question-asking skills, multi-disciplinary collaboration, and peer discussions between students. Our experiments show that simulated virtual students who underwent training with MEDCO not only achieved substantial performance enhancements comparable to those of advanced models, but also demonstrated human-like learning behaviors and improvements, coupled with an increase in the number of learning samples. This work contributes to medical education by introducing a copilot that implements an interactive and collaborative learning approach. It also provides valuable insights into the effectiveness of AI-integrated training paradigms.

\end{abstract}

\section{Introduction}
\label{sec:intro}

Large language models (LLMs)~\cite{achiam2023gpt,touvron2023llama} have transformed various research domains, with their exceptional language processing and understanding skills emerging as generalist intelligence~\cite{sun2023survey}. In medicine and healthcare, LLMs have been explored in numerous areas~\cite{thirunavukarasu2023large,qiu2023large}, from clerical work automation such as writing discharge summaries~\cite{patel2023chatgpt}, to health monitoring such as dietary intake assessment~\cite{lo2024dietary}, and to an entire clinical pathway from patient presentation to treatment involving clinical data interpretation and diagnostic suggestions~\cite{qiu2024application}. With their comprehensive knowledge learned using web-scale data, a promising usage of LLMs is for medical education~\cite{lee2023rise,wu2024embracing}. Globally, the demand for well-trained clinicians has surpassed the capacity to adequatly train the quality medical workforce. Over the past decades, with the development of information engineering and artificial intelligence (AI) technologies, there is a gradual adoption of these technologies within medical education to facilitate training, from online tutoring to AI-enabled assessment. Although recent research~\cite{saab2024capabilities} has shown that LLMs can score high with the accuracy of 91.1\% in medical examinations such as United States Medical Licensing Examination (USMLE), their potentials and effectiveness as copilots to train medical students are yet to be explored. While some work has shared perspectives on using a chatbot such as ChatGPT in medical education~\cite{lee2023rise,wu2024embracing}, a single chatbot can only play a single role at a time while engaging dialogue with a medical student, whereas in real-world medical training, students encounter various patients and clinical cases, and receive supervision and feedback from senior doctors to improve their clinical and patient-facing skills. Throughout the entire training course, the engagement with multiple different individuals such as patients and senior doctors, and a multi-disciplinary team, is essential to the success of transitioning a medical student to be a medical expert. A single chatbot with a single role is limited to deliver such multi-individual and multi-disciplinary training.

More importantly, medical students need to be proficiently trained in the art of asking patients questions. Proficient question-asking skills can lead patients to recall and share more pertinent information, thereby facilitating a more accurate differential diagnosis. Nonetheless, existing AI-assisted educational tools fail to equip medical students with these essential question-asking skills. Furthermore, current educational tools are primarily designed for solitary learning, but medicine at its core is a multi-disciplinary subject, often requiring collaborations from multiple specialties and clinical sectors. Peer discussions are therefore beneficial in helping students reflect and digest medical and clinical knowledge. Current AI-assisted educational tools that do not sufficiently encourage peer discussions and collaborative learning may fall short in instilling the concept of collaboration to the students.

\begin{figure}[tb]
  \centering
  \includegraphics[width=\linewidth]{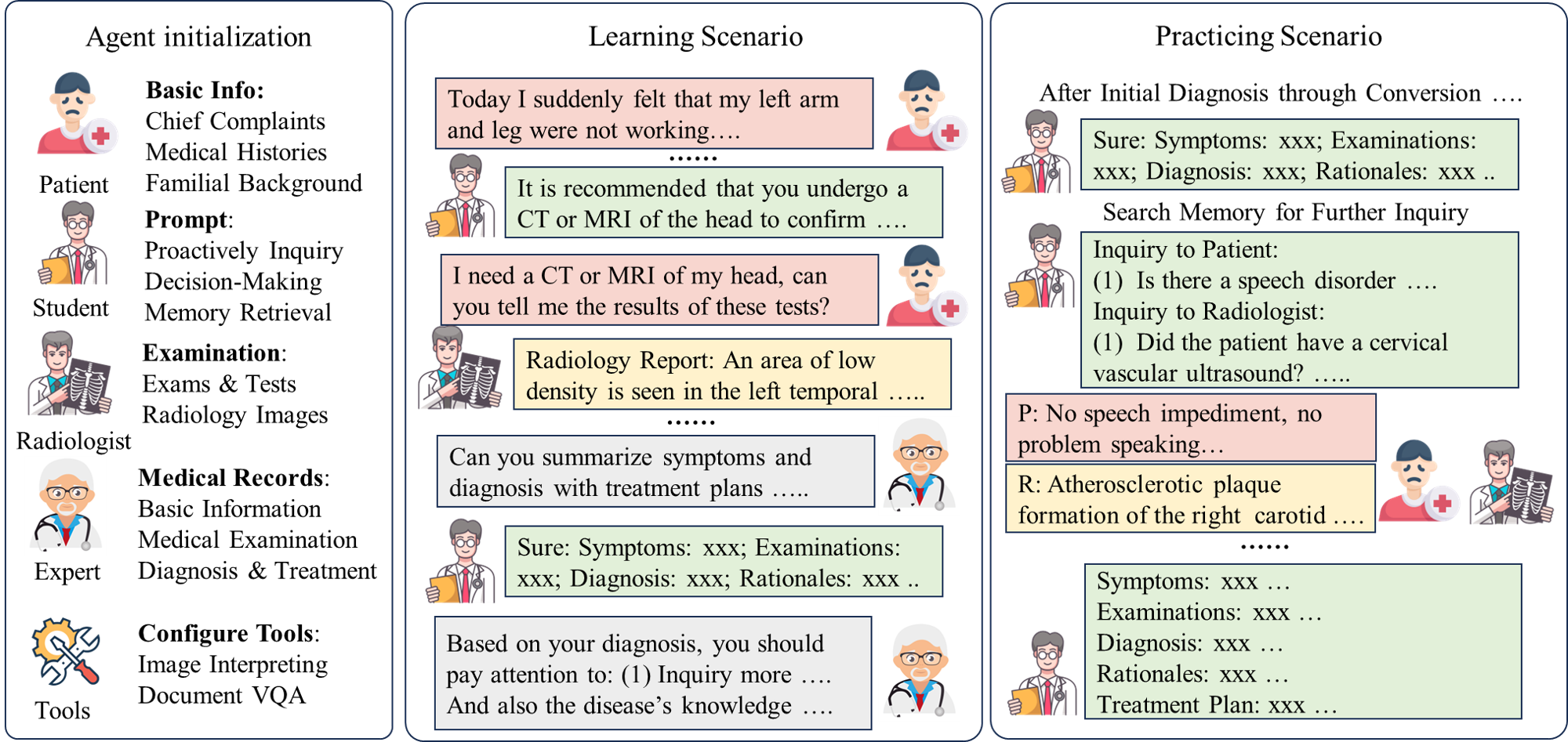}
  \caption{The illustration of the MEDCO framework, which consists of three steps: 1) Initiating different roles and tools; 2) The medical expert evaluates the student's diagnosis and provides feedback, and the student digests and stores the feedback in their learning memory; 3) The student applies the knowledge within memory to improve diagnosis in future practicing scenarios.}
  \label{fig_demostration}
  \vspace{-2em}
\end{figure}

To realize such a training paradigm for AI-enabled medical education, and to overcome the limitations of a single chatbot, we hence propose a multi-agent-based copilot for medical education, based on the conception that a medical student has to have adequate exposure to patient encounters, feedback from senior doctors, multi-departmental collaboration, and peer discussions for the ease of learning. Such an AI-integrated, clinical case-based training differs from conventional reinforced online Q \& A exams by offering more personal and precision medical education. We call our proposed copilot as \textbf{MEDCO}, i.e., \textbf{M}edical \textbf{ED}ucation \textbf{CO}pilots. While there are some LLMs adapted to be a medical specialist (e.g., LLaVa-Med~\cite{li2024llava}, and Med-Gemini~\cite{saab2024capabilities}), and LLM-based agentic frameworks to demonstrate the multi-agent concept in medicine (e.g., Agent Hospital~\cite{li2024agent}), so far there is no agentic system proposed replicating the real-world medical training paradigm and studies its effectiveness. 

The framework of MEDCO is shown in Fig.\ref{fig_demostration}, which includes three stages: agent initialization, learning, and practicing scenarios. In the first step, MEDCO currently involves three agents but has the capacity to incorporate more up to a user's configuration. These three agents are: 1) an agentic patient that can simulate various symptoms and health conditions, and converse with the student to allow the student to practice; 2) an agentic medical expert that can assess the diagnosis of the student and then provide suggestions for improvement and summarize the case-specific knowledge; and 3) an agentic doctor from a separate department to simulate multi-departmental collaboration. The agentic patient and the medical expert are two integral roles within the system, whereas the agentic doctor(s) can be configured as per the requirements. In the current configuration, we set the agentic doctor as a radiologist for interpreting medical images and physical examinations (multi-modality support).

The learning scenario begins with an interactive diagnosis of the student. Based on this diagnosis, the medical expert gives feedback, including suggestions for improvement and relevant disease-related knowledge. In the practicing scenario, after the initial diagnosis, the student can revisit their learned knowledge or suggestions to formulate more differential questions for a final diagnosis. If there are two students, they can discuss their diagnoses together to achieve better results. 

At the current stage, we demonstrate that MEDCO is effective for agentic learning (i.e., students played by LLM models), but we envisage when human users, i.e., actual medical students, interact with MEDCO, they can also effectively improve their medical skills and knowledge. To evaluate the effectiveness of our MEDCO, in addition to the metrics outlined in \cite{fan2024ai}, we introduce new metrics incorporating the ICD-10 (International Classification of Diseases) \cite{world2004international} code for hierarchical performance assessment (from coarse to fine-grained level). Our experiments demonstrate that MEDCO can significantly enhance the performance of a mediocre LLM model, making it comparable to advanced models. For instance, a student initialized with GPT-3.5 can achieve performance levels near or exceeding those of GPT-4o-mini\cite{achiam2023gpt} or Claude3.5-Sonnet\cite{claude35}. Moreover, students initialized with strong models like GPT-4o-mini or Claude3.5-Sonnet also benefit from our agentic learning copilot. During experiments, we observed human-like learning behaviors, indicating that as the agentic students learn more, their performance improves. Additionally, peer discussions between agentic students are shown to enhance the diagnostic performance of proactive students compared to those who learn independently.

In summary, this work makes three key contributions: 1) we introduce MEDCO, a multi-modal, multi-agent copilot designed to enhance medical students' clinical conversion and diagnostic skills; 2) we propose a new hierarchical evaluation metric for assessing diagnostic accuracy at coarse, medium, and fine-grained levels; 3) our findings show that agentic students trained with the copilot exhibit human-like learning behaviors, highlighting its potential for real-world application with actual students.

\section{Related Work}

\subsection{LLMs for Medical Education}

The introduction of ChatGPT has accelerated the exploration of LLMs for medical education~\cite{lee2023rise,wu2024embracing}. The early work of Kung et al.~\cite{kung2023performance} revealed that ChatGPT could achieve a passing score of USMLE. Since then, new generations of LLMs~\cite{singhal2023large,saab2024capabilities} have been benchmarked on USMLE extensively, with the state-of-the-art performance achieving 91.1\% accuracy~\cite{saab2024capabilities}. While the implication of LLMs achieving such a high USMLE score is far-reaching, little research\cite{kung2023performance} has been conducted to integrate LLMs with existing medical educational systems and investigate how effective they are in improving students' learning.

Furthermore, as pointed out in~\cite{qiu2023large}, LLMs in education can be a double-edged sword where learning can be more interactive and responsive with LLMs being assistants, but plagiarism and a decrease of a student's own creativity may also occur and proliferate. We refer readers to~\cite{qiu2023large,abd2023large,wu2024embracing} for more comprehensive discussions about LLMs in medical education

\subsection{LLM-based Agents}

Recently, there has been a trend of shifting from a single LLM chatbot to an agentic LLM framework for task solving, as agents can access tools, plan, reflect, and form collaborations with other agents. Within an educational setting, this agentic paradigm resembles what a human student would experience during the course of learning, i.e., developing reasoning skills, knowing to leverage tools, reflect, and collaborate with others. Hence, an educational system based on multiple agents can be particularly promising. While there are some agentic frameworks, such as ReAct~\cite{yao2022react}, LangChain, CAMEL~\cite{li2023camel}, and AutoGen~\cite{wu2023autogen}, they are mainly proposed for automating workflow. For example, the seminal CAMEL work shows that the workflow of a task can be automated using an agentic user and an agentic assistant in which the agentic user is a proxy of human user to instruct the assistant to fulfill tasks. Recently, agentic systems have been proposed for educational purposes such as simulated classrooms ~\cite{zhang2024simulating,yue2024mathvc}. Lee et al.~\cite{lee23generative} have also proposed to use generative agents to train teachers to help them prepare for the actual in-class teaching.

There are a few studies exploring LLM agents in medical settings, and most of them are unimodal focusing on medical text data only. Agent Hospital~\cite{li2024agent} shows that an agentic doctor can continually improve its diagnosis by merely interacting with agentic patients in a simulated hospital, and can transfer its learned knowledge to real-world cases. MEDAGENTS~\cite{tang2023medagents} demonstrates that multi-round discussions among agents can lead to better results than zero-/few-shot chain-of-thought~\cite{wei2022chain} on medical question answering. Similarly, AI Hospital~\cite{fan2024ai} shows that multi-agent discussions can enhance diagnostic accuracy. Almanac~\cite{zakka2024almanac}, a retrieval-integrated agentic system, shows consistently better performance than plain LLMs in clinical question answering. Nonetheless, the above explorations of agents within medicine are still restricted by the single text modality, whereas medicine is inherently multi-modal.

Recently, MMedAgent~\cite{li2024mmedagent} shows that the medical task-solving capabilities of LLMs can be widened, and their unimodal nature can be expanded to multi-modal, through tool access, such as invoking MedSAM~\cite{ma2024segment} for medical image segmentation and ChatCAD~\cite{zhao2024chatcad+} for medical report generation.

\section{Method}
Fig. \ref{fig_demostration} illustrates the three main steps in our framework: (1) Agent initialization: prompt different LLMs to play different roles using medical information and different instructions; (2) Learning scenario: the agentic medical student generates a diagnostic report according to interactions with the agentic patient and agentic radiologist, which is then evaluated by an agentic medical expert. The student then integrates diagnostic suggestions and case-specific knowledge from the medical expert into its memory; (3) Practicing scenario: after the initial diagnosis for a new clinical case, the agentic medical student revisits relevant suggestions or knowledge from its memory to propose further questions to the patient and radiologist to reach a more accurate final diagnosis. The following subsections elaborate on these three steps.

\subsection{Agent Initialization}
The proposed MEDCO framework currently encompasses four roles: (1) an agentic patient: articulates symptoms, answers questions honestly, participates in examinations, and expresses concerns; (2) an agentic radiologist: interprets various radiological imaging modalities and reports, providing examination results; (3) an agentic student: simulates physicians' investigative and decision-making processes to inquire patients to make diagnoses; (4) an agentic medical expert: evaluates the student's diagnostic process, offering suggestions and relevant knowledge for future inquiries and diagnoses. Given the specific medical knowledge required for these roles, the involved medical records can be categorized into three types.
\begin{itemize}
  \item \textbf{Basic Information}: includes essential patient data such as chief complaints, medical histories, and personal and familial backgrounds, forming the foundation of the patient's medical profile.
  \item \textbf{Medical Examination}: includes detailed assessments in the diagnostic process, encompassing physical exams, auxiliary tests, radiological, and report images (if applicable).
  \item \textbf{Diagnosis and Treatment}: includes diagnostic results, the rationale for diagnoses, and the prescribed treatment plan, reflecting the decision-making and therapeutic elements of patient care.
\end{itemize}
The above three categories of information will be assigned to respective agentic characters via prompts: the agentic patient will have the access to the basic information, the agentic radiologist will be given the medical examination, and the agentic medical expert will use all three types of information to assess the performance of the agentic student.

\textbf{Agentic Patient}: To prompt a language model as a patient, it is vital to emphasize key patient traits and create a thorough personality profile. The patient should express symptoms, answer questions, participate in exams, and voice concerns. Developing a detailed patient persona—covering lifestyle, experiences, traits, and speech—boosts realistic interactions. In \cite{fan2024ai}, the authors use GPT-4 to examine a patient's lifestyle and traits to produce a statement reflecting their condition. Our study follows this method, enabling the model to mimic authentic patient interactions in medical simulations. The prompt for the patient is shown in Table.\ref{prompt_patient}.

\textbf{Agentic Medical Student}: In creating a prompt for an LLM to imitate a medical student, we seek to reflect physicians' investigation and decision-making ability. The prompt encourages interaction with patient agents for detailed symptom assessments, medical exams, and collaboration with radiologists for additional findings. It then assists the student in evaluating treatment options to develop and communicate an optimal plan. Following expert input, the student would summarize symptoms, examinations, diagnostic results, rationales, and treatment plans in detail. During the practicing scenario, this prompt allows the student to recall relevant diseases and diagnostic suggestions based on symptoms, informing further inquiries to achieve a final diagnosis. This process enhances the realism and educational value of simulations, with experiments primarily using GPT-3.5 for initialization considering its weak performance in medical scenarios\cite{fan2024ai}. Our ablation study shows that even students initialized with a strong model also benefit from our educational copilot. The specific prompt for the medical student is shown in Tables \ref{prompt_student1} to \ref{prompt_student5}.

\textbf{Agentic Radiologist}: In the context of simulating a radiologist using an LLM, our approach focuses on encapsulating the multifaceted role of a radiologist within a hospital environment. Our prompt design emphasizes the core competencies of radiological practice, including proficiency in interpreting diverse imaging modalities such as X-rays, CT scans, MRIs, and ultrasounds. Additionally, we instruct the model to respond to specific test requests from patient agents or physicians, providing relevant and accurate examination results based on the patient's simulated condition. In the prompts, we incorporate the Medical Examination section of medical records to ensure the radiologist communicates findings with professional clarity. Additionally, we have developed two tools to interpret radiological and report photos, producing both a radiology report and a textual description, respectively. The prompt for the radiologist is shown in Table \ref{prompt_tool1} and Table \ref{prompt_tool2}. During conversion, the radiologist will also be referred to as the examiner when someone requests the examination results.

\textbf{Agentic Medical Expert}: We developed an advanced prompting strategy to utilize an LLM in medical education to simulate a medical expert. This role aids in assessment, feedback, and educational support during diagnostic scenarios. The expert would evaluate student performance, emphasizing key skills such as patient interaction and clinical reasoning, and identifying strengths and weakness for improvement regarding students' inquires, diagnostic rationales, and treatment plans. Additionally, the model offers case-specific medical knowledge, including definitions, symptoms, and treatment protocols, for students' learning. The relevant prompts for this role are shown in Table \ref{prompt_expert1}, Table \ref{prompt_expert2}, Table \ref{prompt_expert3}, and Table 13 in \cite{fan2024ai}.

\begin{figure}[tb]
  \centering
  \includegraphics[height=7.5cm]{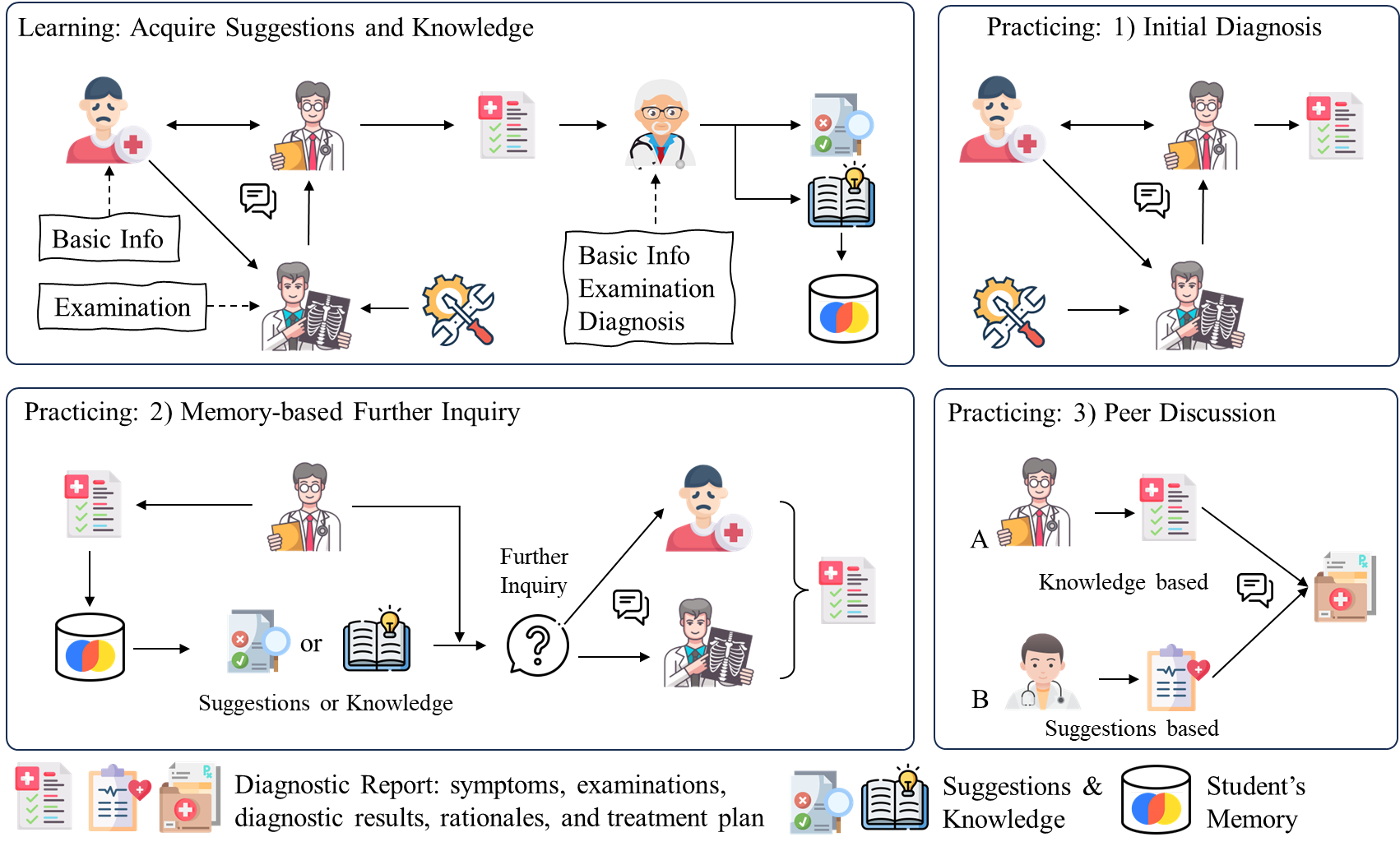}
  \caption{The pipeline of our MEDCO framework from learning to practicing scenario.}
  \label{fig_framework}
\vspace{-2em}
\end{figure}

\subsection{Learning Scenario}

This particular scenario is designed to assist a medical student in acquiring valuable diagnostic insights by engaging in case-based learning, which is shown in Fig.~\ref{fig_framework} and mainly involves three steps:  

\textbf{(1) Initial Diagnosis}: In this essential first step, the student interacts with both the patient and the radiologist in a collaborative setting. The primary goal is to formulate a comprehensive diagnostic report. This report serves to succinctly summarize a range of critical elements, including the patient's reported symptoms, results from various examinations, the diagnosis itself, the reasoning behind the diagnostic conclusions, and the corresponding treatment plans tailored for the patient’s condition. 

\textbf{(2) Assessment}: The medical expert evaluates the report against the complete medical records to provide feedback and suggestions on patient interaction, inquiries, diagnostic hypotheses, clinical reasoning, and treatment plans. Additionally, the expert offers case-specific medical knowledge, including disease definitions, pathogenesis, symptoms, common examinations, and primary treatment protocols, for students' learning. 

\textbf{(3) Learning Feedback}: To simulate the student's learning and practicing process, a key-value memory is implemented to store and retrieve suggestions and case-specific knowledge efficiently. In our setting, saving information to memory is equivalent to a human student's acquisition of this information through learning. The detailed designs of our memory mechanism can be found in section \ref{sec_memory_design} in the appendix. 

\subsection{Practicing Scenario}
The practicing scenario aims to assess the diagnostic performance improvement of the medical student after being trained by our copilot, as illustrated in Fig.~\ref{fig_framework}:

\textbf{(1) Initial Diagnosis}: Similar to step (1) in the learning scenario, this step involves interactive communication with the patient and radiologist to formulate a preliminary diagnostic report.

\textbf{(2) Rethinking and Recalling}: Using the summarized symptoms from the initial diagnosis, the agentic student revisits their learned knowledge of diseases or suggestions to propose differential questions for the patient and radiologist by following the prompts in the appendix (Table \ref{prompt_student3} and Table \ref{prompt_student4}, respectively).

To recall knowledge, the agentic student first retrieves relevant diseases from the memory's symptom-store and then accesses disease-related information from the disease-store. Similarly, the student would utilize patient IDs from these relevant diseases to gather suggestions from the case-store in memory. 

\textbf{(3) Further Inquiry and Diagnosis}: Based on the additional questions, the agentic student inquires the patient and radiologist to gather more differential information for the final diagnosis. Table \ref{prompt_student2} shows the prompt for this inquiry. 

\textbf{(4) Peer Discussion}: In scenarios with two agentic students, each retrieves relevant suggestions and knowledge respectively in step (2). They then engage in a discussion about their diagnosis to improve themselves, where the prompt is outlined in Table \ref{prompt_student5}. This collaboration enhances clinical reasoning and decision-making by integrating their insights.

Figures \ref{fig_knowledge_case_chinese} and \ref{fig_knowledge_case_english} illustrate an interactive diagnosis case that demonstrates the student retrieving knowledge from memory based on symptoms to enhance the final diagnosis.

\section{Experiments}

\subsection{Dataset}

In this study, we used the MVME dataset \cite{fan2024ai} of Chinese medical records, which contains 506 high-quality medical cases (text-only) across various specialties, e.g., surgery, internal medicine, and obstetrics and gynecology. More details about MVME can be found in~\cite{fan2024ai}. We randomly divided the dataset into training (259 cases) and testing (247 cases) sets at the department level for learning and practicing scenarios, respectively.

To validate our framework's multi-modal capability, we selected the entire test set of 16 neurological cases from the department of internal medicine and gathered their corresponding radiological images or report photos from the original websites. Two example images are shown in Fig.\ref{fig_multimodal_data} in the appendix. During the dialogue, the radiologist interprets the images of the current patient into textual descriptions, such as a radiology report or summary of existing reports, when requested to provide such information by the agentic patient or student.

\subsection{Evaluation Metrics}

We assess MEDCO using three evaluation metrics from both qualitative to quantitative aspects. More implementation details can be found at the section \ref{sec_metrics} in the appendix

\textbf{HDE (Holistic Diagnostic Evaluation)}: the rating (1$\sim$4 points) of the expert regarding the student's report, including five sections: symptoms, medical examination, diagnostic results, rationales, and treatment plans.

\textbf{SEMA (Semantic Embedding-based Matching Assessment)}: First, retrieving the top-10 relevant ICD terminologies of the student's diagnosis and the ground truth (the medical record), respectively, and then compute their extracted disease entities (\#), precision(P), recall(R), and F1-score(F1) as metrics.

\textbf{CASCADE (Coarse And Specific Code Assessment for Diagnostic Evaluation)}: Similar to the SEMA, we first retrieve the top-1 relevant ICD terminology and then compute the accuracy at three ICD-10 levels: coarse, medium, and fine-grained levels.

\subsection{Implementation Details}
In MEDCO, distinct models are assigned to various roles: GPT-3.5\cite{achiam2023gpt} acts as the patient, while Claude-3.5-Sonnet-20240620\cite{claude35} serves as both the radiologist and medical expert, considering its excellence in biology subjects\cite{huang2024olympicarena}. The implementation of the student's memory and ICD-10 metrics (SEMA \& CASCADE) uses Chromadb with OpenAI embedding for efficient storage and rapid queries.

To simulate the agentic radiologist's interpretation of various imaging modalities, such as radiological images and report photos, we design two tools: 1) the Radiology tool for interpreting radiological images, and 2) the ReportVQA tool for describing the report photos. More details can be found at the section \ref{sec_tool} in the appendix.

In addition, the diagnosis of two doctors (GPT-4 and GPT-3.5) powered by the multi-agent system in \cite{fan2024ai} serves as the upper bound in our experiments.

\subsection{Performance of the Agentic Student}


Table \ref{student_performance} presents evaluations from the medical expert regarding a student's report, indicating significant performance gains after the learning scenario.

The Claude3.5-Sonnet model scored the highest average score of 2.283 (±0.328), while the student (GPT-3.5) achieved an average of 1.965 (±0.336). After learning, the agentic student's overall performance improved, with scores rising to 2.169 (±0.337) through recalling learned knowledge and 2.122 (±0.341) through revisiting suggestions. Peer discussions yielded the best score of 2.299 (±0.393), outperforming both the Claude3.5-Sonnet and the upper-bound 2-agent benchmarks. Particularly, the Medical Examination and Diagnostic Rationales sections achieved significant gains, with the Medical Examination score rising from 1.785 to 2.575 following peer discussions, and Diagnostic Rationales improving from 1.879 to 2.158.

\begin{table}
\vspace{-1em}
\scriptsize
\caption{HDE results of the agentic student with and without the training of our copilot across the entire test set.}
\label{student_performance}
\centering
\begin{tabular}{l|ccccc|c}
\hline
 & Symptom & Medical & Diagnostic & Diagnostic & Treatment & Avg\\
 & & Examination & Results & Rationales & Plan & (std)\\
\hline
 & \multicolumn{5}{c}{Student} \\
\hline
GPT-3.5 & 2.595  & 1.785  & 1.960  & 1.879 & 1.607  & 1.965(0.336)\\
GPT-4o-mini & 2.688 & 1.980 & 2.134 & 1.931 & 1.628 & 2.072(0.349)\\
Claude3.5-Sonnet & \textbf{2.895}  & 2.113 & \textbf{2.247} & \textbf{2.243} & \textbf{1.919} & 2.283(0.328)\\
\hline
 & \multicolumn{5}{c}{Student(GPT-3.5) + Learning Scenario}  \\
\hline
w/ knowledge & 2.696 & 2.296 & 2.113 & 2.085 & 1.656 & 2.169(0.337)\\ 
\hline
w/ suggestions & 2.662 & 2.263 & 2.061 & 2.008 & 1.619 & 2.122(0.341)\\
\hline
w/ discussion & 2.866 & \textbf{2.575} & 2.178 & 2.158 & 1.717 & \textbf{2.299(0.393)}\\

\hline
 & \multicolumn{5}{c}{Upper bound} \\
\hline
2 Agents\cite{fan2024ai}  & 2.741 & 2.259 & 2.227 & 2.130 & 1.870 & 2.245(0.283)\\
\hline
\end{tabular}
\vspace{-2em}
\end{table}

Table~\ref{student_f1} illustrates ICD-10 metrics (SEMA and CASCADE), showing notable enhancements in diagnostic performance for the agentic student after learning. In terms of SEMA metrics, the student (GPT-3.5) achieved remarkable increases in recall (29.72) and F1-score (36.04) through peer discussions, exceeding the untrained student (R: 17.95, F1: 26.01). Claude3.5-Sonnet had the highest precision (55.13).

CASCADE metrics showed improvement across the board, with the student (GPT-3.5) achieving top accuracy levels (46.67$\%$ and 18.33$\%$ for coarse and medium, respectively). While peer discussions were strong in the recall, they did not enhance accuracy at coarse and medium levels, likely due to more false positives from higher entity extraction (2.59).

Despite the Claude3.5-Sonnet consistently achieving the best accuracy (48.26$\%$, 22.23$\%$, and 12.23$\%$ for coarse, medium, and fine-grained levels, respectively), the trained student (GPT-3.5) demonstrated comparable performance, further demonstrating our framework's effectiveness.

\begin{table}
\vspace{-1em}
\caption{The ICD-10 related results (SEMA and CASCADE score) for the agentic student with and without the training of our copilot across the entire test set.}
\label{student_f1}
\centering
\begin{tabular}{l|c|ccc|ccc}
\hline
 & \# & R$\uparrow$ & P$\uparrow$ & F1$\uparrow$ & Coarse($\%$) $\uparrow$ & Medium ($\%$) $\uparrow$ & Fine ($\%$) $\uparrow$\\
\hline
\multicolumn{8}{c}{Student} \\
\hline
GPT-3.5 & 1.52 & 17.95 & 47.20 & 26.01  &43.72 & 17.37 & 8.31  \\
GPT-4o-mini & 1.82 & 23.43 & 51.45 & 32.20 & 45.61 & 19.17 & 11.48 \\
Claude3.5-Sonnet & 1.58 & 21.81 & \textbf{55.13} & 31.25  & \textbf{48.26} & \textbf{22.23} & \textbf{12.23} \\
\hline
\multicolumn{8}{c}{Student(GPT-3.5) + Learning Scenario} \\
\hline
w/ knowledge & 1.79 & 22.31 & 49.77 & 30.81& 46.67 & 18.33 & 9.91 \\
w/ suggestions & 1.91 & 23.12 & 48.41 & 31.30 & 44.76 & 17.93 & 10.53 \\
w/ discussion & 2.59 & 29.72 & 45.78 & 36.04 & 44.31 & 17.27 & 10.00 \\ 
\hline
\multicolumn{8}{c}{Upper bound} \\
\hline
2 Agents\cite{fan2024ai} & 2.62 & \textbf{32.66} & 49.69 & \textbf{39.41} & 46.08 & 19.81 & 11.42 \\
\hline
\end{tabular}
\vspace{-2em}
\end{table}

\begin{table}
\scriptsize
\vspace{-1em}
\caption{The HDE results of the agentic student role-played by a strong model.}
\label{strong_model_perfmance}
\centering
\begin{tabular}{l|ccccc|c}
\hline
 & Symptom & Medical & Diagnostic & Diagnostic & Treatment & Avg\\
 & & Examination & Results & Rationales & Plan & (std)\\
\hline
 & \multicolumn{5}{c}{Student} \\
\hline
Claude3.5-Sonnet & 2.895  & 2.113 & 2.247 & 2.243 & 1.919  & 2.283(0.328)\\
\hline
&\multicolumn{5}{c}{Student(Claude3.5-Sonnet) + Learning Scenario} \\
\hline
w/ knowledge & 3.045 & 2.915 & 2.425 & 2.457 & 2.089 & 2.586(0.358)\\
w/ suggestion & 3.065 & 3.138 & 2.453 & 2.603 & 2.206 & 2.693(0.128)\\
w/ discussion & 3.121 & 3.032 & 2.453 & 2.632 & 2.190 & 2.686(0.350)\\
\hline
 & \multicolumn{5}{c}{Upper bound} \\
\hline
2 Agents\cite{fan2024ai}  & 2.741 & 2.259 & 2.227 & 2.130 & 1.870 & 2.245(0.283)\\
\hline
\end{tabular}
\vspace{-2.0em}
\end{table}

\subsection{Performance Gain for the Students Played by Strong Models}\label{sec_strong} 

This subsection explores how the copilot improves performance in students played by strong language models, exemplified by Claude3.5-Sonnet. As shown in Tables \ref{strong_model_perfmance} and \ref{strong_model_f1}, MEDCO significantly enhances agentic students' clinical consultation and diagnostic abilities, demonstrating its effectiveness across different model architectures.

\begin{table}
\vspace{-1em}
\caption{The ICD-10 related assessment (SEMA and CASCADE score) for the agentic student played by a strong model (Claude3.5-Sonnet)}
\label{strong_model_f1}
\centering
\begin{tabular}{l|c|ccc|ccc}
\hline
 & \# & R$\uparrow$ & P$\uparrow$ & F1$\uparrow$ & Coarse($\%$) $\uparrow$ & Medium($\%$) $\uparrow$ & Fine($\%$) $\uparrow$\\
\hline
\multicolumn{8}{c}{Student} \\
\hline
Claude3.5-Sonnet & 1.58 & 21.81 & 55.13 & 31.25  & 48.26 &22.23 & 12.23 \\
\hline
\multicolumn{8}{c}{Student(Claude3.5-Sonnet) + Learning Scenario} \\
\hline
w/ knowledge & 1.97 & 27.79 & 56.26 & 37.20 & 48.16 & 21.82 & 12.58 \\
w/ suggestions & 1.99 & 29.01 & \textbf{58.13} &38.70 & \textbf{49.55} & \textbf{23.38} & \textbf{13.84} \\
w/ both & 2.25 & 32.56 & 57.73 & 41.63 & 48.14 & 22.49 & 12.96 \\
\hline
\multicolumn{8}{c}{Upper bound} \\
\hline
2 Agents\cite{fan2024ai} & 2.62 & \textbf{32.66} & 49.69 & \textbf{39.41} & 46.08 & 19.81 & 11.42 \\
\hline
\end{tabular}
\end{table}

Table \ref{strong_model_perfmance} reveals that the untrained agentic student scores an average of 2.283 (±0.328) in five diagnostic aspects. With copilot training, scores rise: students using recalled knowledge could reach the average of 2.586 (±0.358), while those using recalled suggestions could reach 2.693 (±0.128). Peer discussions yield the highest score of 2.686 (±0.350), surpassing the 2 Agents, the average of 2.245 (±0.283). Individual performance improvements are also consistent across dimensions. For example, the Medical Examination score improves from 2.113 (±0.089) to 3.138 (±0.093) when recalling learned suggestions.

Table \ref{strong_model_f1} supports these findings with SEMA and CASCADE metrics, showing improved recall, precision, and F1 scores, with peer discussions achieving a high F1 score of 41.63. Trained agentic students show better accuracy in ICD-10 classifications, particularly those using recalled suggestions, achieving higher scores than untrained peers.

These results indicate that MEDCO can significantly boost the performance of advanced language models like Claude3.5-Sonnet in clinical contexts, showing improvements across various metrics.

\subsection{Learning Curve of the Agentic Student}\label{sec_learning_curve}

\begin{figure}[tb]
  \centering
  \includegraphics[width=\linewidth]{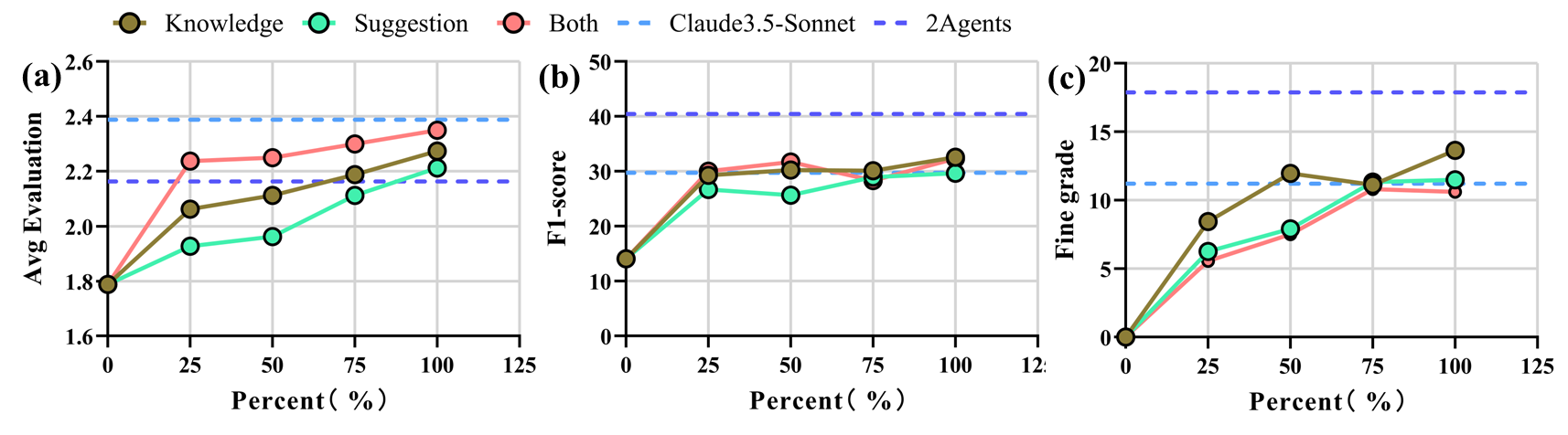}
  \caption{The learning curve of an agentic student using rethinking strategies (knowledge and suggestion) or peer discussion at various retrieval ranges (as percentages) in the practicing scenario, where the performance of Claude3.5-Sonnet and 2 Agents serve as reference benchmarks.}
  \label{fig_learning_curve}
  \vspace{-2em}
\end{figure}

This section examines the learning curve of the agentic student (GPT-3.5) as the number of training cases increases. We focus on Neurology cases from Internal Medicine with 16 training and 16 test cases. In the practicing scenario, when recalling learned experiences or participating in peer discussion, the agentic student's memory retrieval range is limited to 0$\%$ (no training), 25$\%$ (4 cases), 50$\%$ (8 cases), 75$\%$ (12 cases), and 100$\%$ (all 16 training cases), to see how different quantities of training samples could affect diagnostic performance.

In Fig.\ref{fig_learning_curve}(b), the F1-score rises quickly, stabilizing between the 25$\%$ and 100$\%$ ranges, indicating that even limited training can enhance precision and recall. Fine-grained level accuracy in Fig.~\ref{fig_learning_curve}(c) dramatically increases from 0$\%$ to 75$\%$, with the agentic student that rethinks knowledge often surpassing Claude3.5-Sonnet. More detailed results with all metrics are available in Tables \ref{curve_performance} and \ref{curve_f1} in the appendix.

Overall, MEDCO significantly improves agentic students' clinical skills, showing human-like learning behaviors. This consistent performance suggests its potential for real-world application in hospitals, aiding students' educational journeys.

\subsection{Multi-Modal Support}\label{sec_multimodal}

The MVME dataset contains only text, so we collected corresponding images to test our framework’s feasibility with multi-modality. We chose 16 Neurology cases from the Department of Internal Medicine for this study. In interactive diagnosis, radiologists convert visual data such as radiological images or medical report photos into textual descriptions for student comprehension. An example of this is shown in Fig.~\ref{fig_multimodal_case}.

\begin{figure}[tb]
  \centering
  \includegraphics[width=\linewidth]{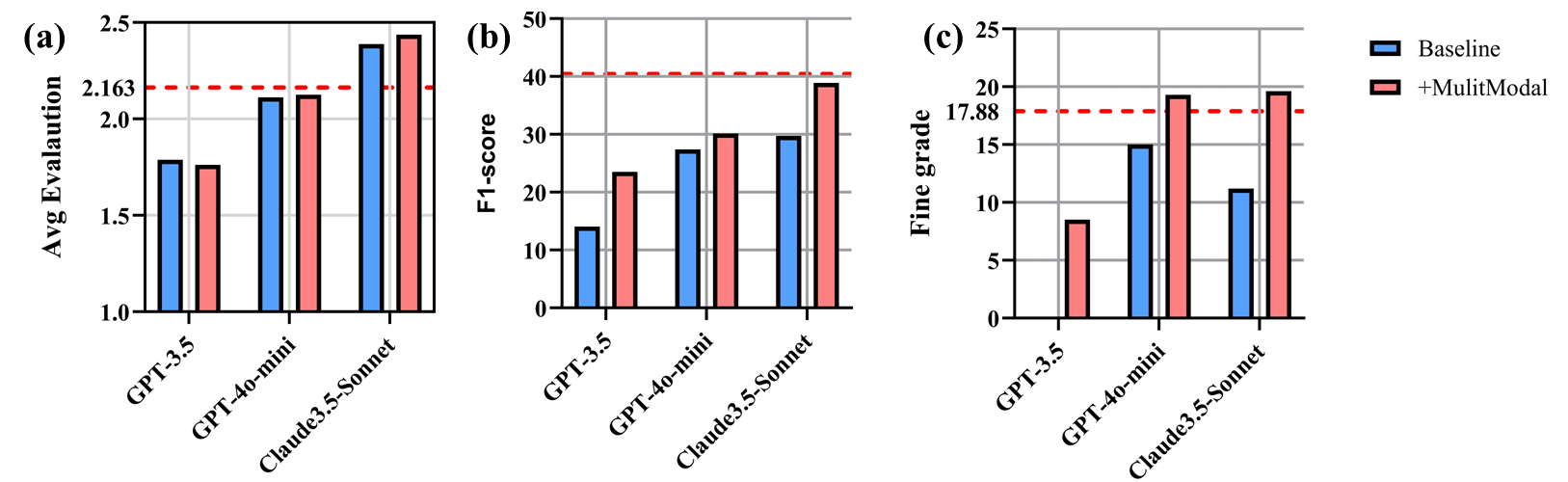}
  \caption{The influence of multi-modalities on the performance of the agentic student when initialized by different LLMs, where the red dashed line denotes the results of 2 Agents\cite{fan2024ai}, serving as the reference.}
  \label{fig_multimodal}
\vspace{-2em}
\end{figure}

We utilized GPT-4o-mini, Claude3.5-Sonnet, and GPT-3.5 to initialize the patient role and evaluate the impact of visual data on them. Results in Fig.~\ref{fig_multimodal} show average scores from expert evaluations, F1-scores (SEMA), and fine-grained level accuracy (CASCADE), with additional metrics in Tables \ref{multimodal_performance} and \ref{multimodal_f1}. From the results, Fig.~\ref{fig_multimodal}(a) reveals that multi-modal input benefits students initialized with GPT-4o-mini and Claude3.5-Sonnet, particularly the latter, which achieves the highest results, surpassing 2 Agents\cite{fan2024ai}.

The F1-score in Fig.~\ref{fig_multimodal}(b) indicates a strong positive effect of multi-modal input for all variants. the agentic student(GPT-3.5) shows the highest relative improvement, while Claude3.5-Sonnet variant achieves the top absolute F1-score, approaching the 2 Agents benchmark. Additionally, fine-grained level accuracy in Fig.~\ref{fig_multimodal}(c) illustrates significant enhancements from multi-modal input, suggesting visual data improves detailed fine-grade diagnosis. Notably, multi-modal input significantly boosts GPT-4o-mini and Claude3.5-Sonnet student performance to surpass the reference.

These results suggest that the multi-modal copilot could aid students’ understanding of complex diagnoses, particularly for fine-grade assessments, by combining visual and textual information to foster a more engaging learning environment.

\section{Discussion}

Case-based learning is the core of our MEDCO framework for AI-enabled medical education. It allows a student to actively interact with a virtual agentic patient, auxiliary doctor, and medical expert for diagnosing a clinical case. This differs from a single-agent chatbot in that instead of one AI character is available, multiple virtual characters are simulated in MEDCO and available for discussion at the request of the student user, replicating real-world clinical settings where medical consultation and diagnosis can be multidisplinary and require expertise from senior doctors as well as active engagement, cooperation from the patient. While we have validated MEDCO from various dimensions, the student currently is only simulated and played by an LLM such as GPT-3.5. Although we have used LLMs of different versions to simulate students with different learning capabilities, and a memory mechanism to represent a student's learning and the acquisition of new knowledge process, there is clearly a gap between the real and virtual students in learning, and hence the effectiveness of MEDCO for helping human students in learning medical knowledge requires further investigation. A larger-scale multi-modal collaborative dataset on par with industrial datasets in other domains~\cite{yu2023v2x} could further unlock the potentials of the MEDCO system. In current implementation, the student receives only textual feedback from the medical expert. Future work can enable the expert to provide quintessential medical imaging examples of a learning case to offer multi-modal feedback to improve the student's diagnostic skill and broaden their knowledge. Similarly, the capabilities of agentic doctors for multi-departmental collaboration can also be expanded. For example, the agentic radiologist can also be granted access to foundation AI models~\cite{christensen2024vision,chen2024towards,qiu2023visionfm,wei2024visionclip,lam2024foundation} or specialized AI models~\cite{ma2024segment,shi2023generalist} as their tools, e.g., using a segmentation model to segment and highlight lesions which could potentially ease the learning of the student.

\section{Conclusion}

We have introduced MEDCO, an innovative multi-agent copilot system for medical education. MEDCO enables medical students to interact dynamically with an agentic patient, specialized doctor, and medical expert to learn about the diagnosis and treatment of various diseases. Our findings indicate that MEDCO can effectively adapt a generalist model for medical specialization. This adaptation may also occur for human students using MEDCO, as it provides targeted feedback tailored to individual learning cases and simulates diverse patient encounters and multidisciplinary collaborations to enhance the learning experience. Furthermore, this AI-enabled copilot system has the potential for application beyond medicine, benefiting broader educational contexts.

\bibliographystyle{splncs04}
\bibliography{main}

\clearpage
\appendix

\section{Acknowledgements}
Thanks for free icons from Flaticon.com\footnote{Flaticon: \url{https://www.flaticon.com/}}.

\section{Prompts}
This section lists the used prompts for the involved roles in our study, where `{xx}' denotes the external inputs in prompts or conversion.

\begin{table*}[ht!]
\scriptsize
    \centering
    \begin{tabular}{lccc}
    \toprule
      Prompt  &Medical Role &Function  \\
    \midrule
       Table\ref{prompt_patient}  &Patient  & Chat with Student and Radiologist \\
       Table\ref{prompt_student1} & Medical Student & Interactive Clinical Diagnosis \\
       Table\ref{prompt_student2} & Medical Student & System message for further inquiry conversion \\
       Table\ref{prompt_student3} & Medical Student & Raise questions for patient based on knowledge \\
       Table\ref{prompt_student4} & Medical Student & Raise questions for radiologist based on knowledge \\
       Table\ref{prompt_student5} & Medical Student & Peer discussion for knowledge and suggestions based diagnosis \\
       Table\ref{prompt_radiologist1} & Radiologist & Process Examination Request  \\
       Table\ref{prompt_radiologist2} & Radiologist & Call tools to interpret images  \\
       Table\ref{prompt_tool1} & Radiology Tool & Interpret radiology images as report  \\
       Table\ref{prompt_tool2} & ReportVQA Tool & Interpret existing report photos as textual descriptions \\
       Table\ref{prompt_expert1} & Medical Expert & Ask the student to summarize the diagnosis \\
       Table\ref{prompt_expert2} & Medical Expert & Assess the diagnosis report of the student\\
       Table\ref{prompt_expert3} & Medical Expert & Summarize the case-specific knowledge \\
       Table 13\cite{fan2024ai} & Medical Expert & Evaluate the student's diagnosis with five criterions\\
    \bottomrule
    \end{tabular}
    \caption{Prompts of different medical roles and the corresponding function in our study.}
    \label{tab:prompt_summary}
    \vspace{-2em}
\end{table*}

\section{Memory Design}\label{sec_memory_design}
Using the dialogue case in the appendix (Fig. \ref{fig_learning_case_chinese} and Fig. \ref{fig_learning_case_english}) as an example, our memory design includes three components to store the feedback of the medical expert in the dialog:
\begin{itemize}
    \item Case-store: the key represents the patient ID, and the value contains the expert's diagnosis suggestion for that case;
    \item Disease-store: the key identifies the diseases in the original medical record (the ground truth), such as cerebral infarction, hypertension, and hyperlipidemia, with the value corresponding to the disease-related knowledge from the expert;
    \item Symptom-store: the key indicates the patient's symptoms in the original medical record (the ground truth), like sudden onset of unclear speech and impaired mobility of the left limbs, with the corresponding value containing the diagnosis results for the case and patient ID.
\end{itemize}

\begin{figure}[tb]
  \centering
  \includegraphics[width=\linewidth]{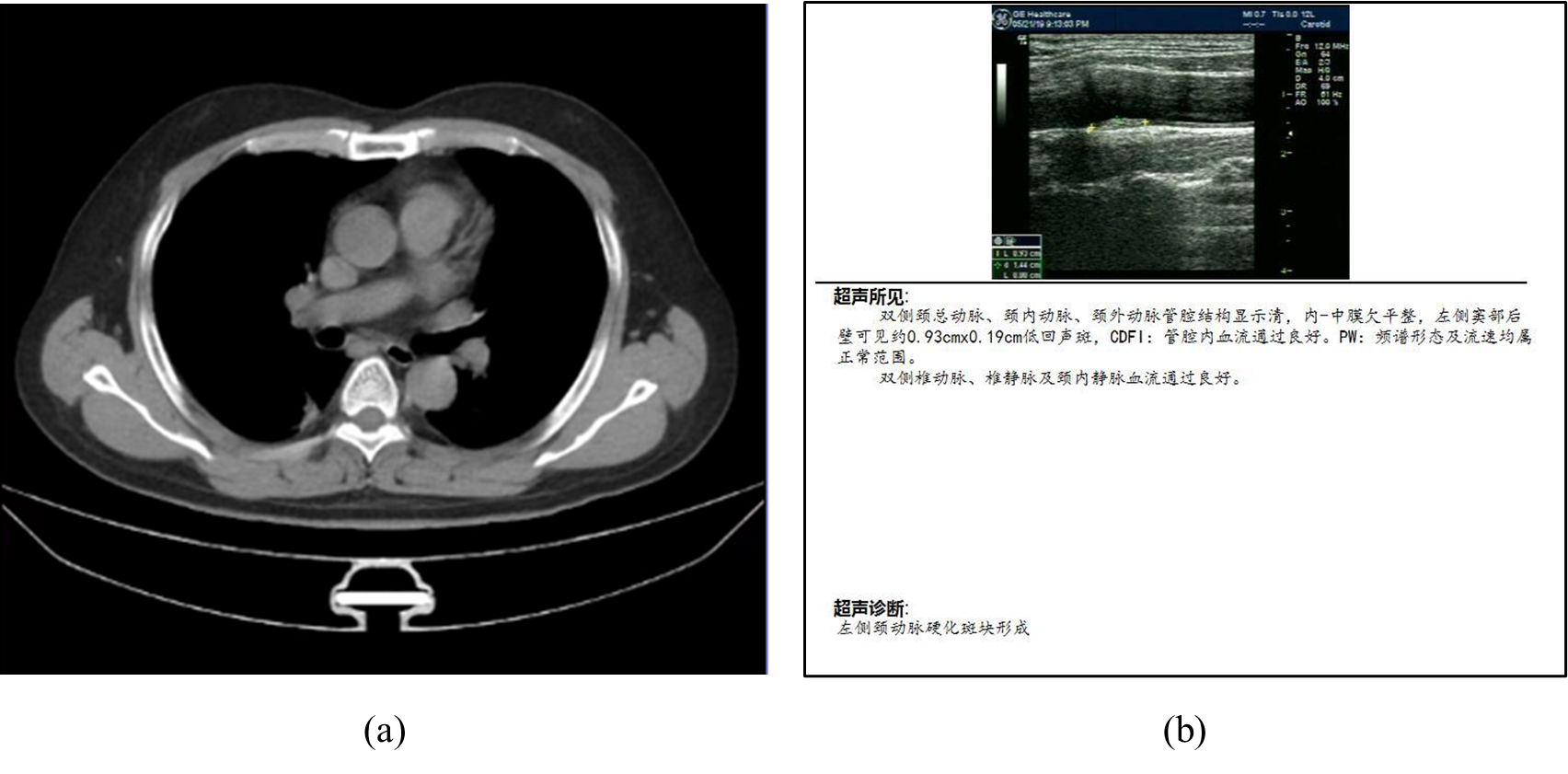}
  \caption{The examples to show our collected visual data: (a) The Chest CT image from patient 1170; (b) The neck ultrasound report photo (Chinese text) from patient 1310.}
  \label{fig_multimodal_data}
\end{figure}

\section{Metrics}\label{sec_metrics}
In our experiments, we design the following three metrics to validate the effectiveness of our proposed copilot: MEDCO:

\textbf{HDE (Holistic Diagnostic Evaluation)}: This qualitative evaluation method\cite{fan2024ai} focuses on holistically analyzing the entire diagnosis process across five specific dimensions: (1) symptoms, (2) medical examination, (3) diagnostic results, (4) diagnostic rationales, and (5) treatment plan. Using medical records as a reference, the medical expert rates the student's performance on a scale of 1 to 4, from poorest to excellent, based on responses to a predefined criterion in the prompt.

\textbf{SEMA (Semantic Embedding-based Matching Assessment)}: We further use the International Classification of Diseases (ICD-10)\cite{world2004international} as our reference standard. We extract disease entities from both the diagnostic results and medical record references, utilizing inner-product similarity to link these entities to ICD-10 terminology, thereby creating normalized disease sets. Unlike the original fuzzy matching process\cite{fan2024ai}, which relies solely on text similarity and may overlook semantically similar yet textually distinct samples, we implement the OpenAI embedding function (`text-embedding-3-large') to project queries and terminology into feature space, allowing for the calculation of inner products to identify the top-10 results. We then evaluate the diagnostic agreement between student diagnoses and the ground truth from medical records, reporting the number of extracted disease entities (\#), set-level precision (P), recall (R), and F1-score (F). This method ensures a standardized and comprehensive assessment of diagnostic accuracy.

\textbf{CASCADE (Coarse And Specific Code Assessment for Diagnostic Evaluation)}: The ICD-10\cite{world2004international} coding system uses a hierarchical structure to classify diseases and health conditions with increasing specificity. It comprises 22 chapters, each identified by a letter (A-Z, excluding U), representing major disease categories. Within these chapters, conditions are categorized by three-character codes, starting with the chapter letter followed by two digits. Many three-character categories are further subdivided with a fourth character after a decimal, and fifth or sixth characters can also be added for greater specificity. For example, `A16.202' indicates `Chapter I: Certain infectious and parasitic diseases` through `A', `Respiratory tuberculosis, not confirmed bacteriologically or histologically' with `A16', and provides further specificity with `A16.202', which denotes `Tuberculosis'. We link disease entities to the ICD-10 terminology and hierarchically assess accuracy at three levels: coarse (A), medium (A16), and fine grade (A16.202).

\section{Tools}\label{sec_tool}
To simulate the agentic radiologist's interpretation of various imaging modalities, such as radiological images and report photos, we offer the following tools:
\begin{itemize}
  \item Radiology tool: This tool interprets radiological images—like X-rays, CT scans, and MRIs—into radiology reports that highlight key findings and impressions. It is powered by the GPT-4o-mini using the specialized prompt in Table \ref{prompt_tool1}.
  \item Report VQA (vision question-answer) tool: This tool translates existing medical report records into textual descriptions, utilizing Claude-3.5-Sonnet-20240620 with the prompt in Table \ref{prompt_tool2}.
\end{itemize}
These tools convert visual data into text, mimicking how radiologists interpret patients' imaging examinations to generate reports or answer physician questions. Except for current API-based tools, this framework supports more complex options from Huggingface\footnote{Huggingface: \url{https://huggingface.co/}}, ModelScope\footnote{ModelScope:\url{https://www.modelscope.cn/home}}, or self-hosted models.

\section{Detailed results}
This section contains the detailed results of the experiments of learning curve exploration in the section \ref{sec_learning_curve} (Table \ref{curve_performance} and Table \ref{curve_f1}) and multi-modality validation in the section.\ref{sec_multimodal} (Table \ref{multimodal_performance} and Table \ref{multimodal_f1}).

\begin{table}
\scriptsize
\caption{The HDE results in the Neurology cases of the department of \textbf{Internal Medicine} with different retrieve ranges (percentages) in the practicing scenario.} 
\label{curve_performance}
\centering
\begin{tabular}{l|ccccc|c}
\hline
 & Symptom & Medical & Diagnostic & Diagnostic & Treatment & Avg\\
 & & Examination & Results & Rationales & Plan &(std)\\
\hline
 & \multicolumn{5}{c}{Student} \\
\hline
GPT-3.5 & 2.562 (0.250) & 1.438 (0.250) & 1.688 (0.312) & 1.625 (0.312) & 1.625 (0.312) & 1.788\\
GPT-4o-mini & 2.688(0.250) & 2.000(0.438) & 2.188(0.500) & 1.875(0.312) & 1.812(0.375) & 2.112\\
Claude3.5-Sonnet & \textbf{2.875} (0.188) & 2.375 (0.375) & \textbf{2.375} (0.500) & \textbf{2.188} (0.438) & \textbf{2.125} (0.438) & 2.388\\
\hline
 & \multicolumn{5}{c}{Student(GPT-3.5) + Learning Scenario} \\
\hline
w/ knowledge & 2.812(0.188) & 2.562(0.375) & 2.188(0.375) & 2.000(0.312) & 1.812(0.375) & 2.274\\
same but know 75$\%$ & 2.688(0.250) & 2.438(0.375) & 2.188(0.375) & 1.938(0.312) & 1.688(0.312)&2.188 \\
same but know 50$\%$ & 2.562(0.312) & 2.438(0.438) & 2.062(0.375) & 1.812(0.312) & 1.688(0.312) &2.112\\
same but know 25$\%$ & 2.562(0.312) & 2.250(0.438) & 2.062(0.312) & 1.812(0.250) & 1.625(0.312) &2.062\\
\hline
w/ suggestion & 2.688(0.250) & 2.438(0.312) & 2.312(0.375) & 2.062(0.250) & 1.562(0.312) &2.212\\
same but know 75$\%$ & 2.688(0.250) & 2.312(0.375) & 2.062(0.438) & 1.938(0.312) & 1.562(0.250)&2.112 \\
same but know 50$\%$ & 2.562(0.250) & 2.125(0.375) & 1.812(0.375) & 1.812(0.375) & 1.500(0.250) &1.962\\
same but know 25$\%$ & 2.625(0.250) & 2.125(0.312) & 1.812(0.375) & 1.512(0.375) & 1.562(0.312) &1.927\\
\hline
w/ both & 2.812(0.188) & 2.750(0.312) & 2.188(0.375) & 2.188(0.375) & 1.812(0.375) &2.350\\
same but know 75$\%$ & 2.812(0.188) & 2.688(0.312) & 2.062(0.375) & 2.125(0.312) & 1.812(0.312) &2.300\\
same but know 50$\%$ & 2.688(0.250) & 2.688(0.375) & 2.062(0.375) &2.000(0.375) & 1.812(0.312) &2.250\\
same but know 25$\%$ &2.625(0.312) & 2.562(0.375) & 2.000(0.375) & 2.062(0.312) & 1.938(0.375) &2.237\\
\hline
 & \multicolumn{5}{c}{Upper bound} \\
\hline
2 Agents\cite{fan2024ai} & 2.688(0.250) & 2.125(0.250) & 2.312(0.500) & 1.938(0.375) & 1.750(0.250) &2.163\\
\hline
\end{tabular}
\end{table}

\begin{table}
\caption{The results of the ICD-10 related metrics (SEMA and CASCADE) on Neurology cases in the department of \textbf{Internal Medicine Neurology} with different retrieve ranges (percentages) in the practicing scenario.  }
\label{curve_f1}
\centering
\begin{tabular}{l|c|ccc|ccc}
\hline
 & \# & R$\uparrow$ & P$\uparrow$ & F1$\uparrow$ & Coarse($\%$) $\uparrow$ & Medium($\%$) $\uparrow$ & Fine($\%$) $\uparrow$\\
\hline
\multicolumn{8}{c}{Student} \\
\hline
GPT-3.5 & 1.44 & 10.42 & 21.74 & 14.08  &43.86 & 1.39 & 0.00  \\
GPT-4o-mini & 1.56 & 20.83 & 40.00 & 27.40 & 52.95 & 29.60 & 15.02 \\
Claude3.5-Sonnet & 1.62 & 22.92 & \textbf{42.31} & \textbf{29.73}  & 62.32 & \textbf{34.98} & \textbf{11.20} \\
\hline
\multicolumn{8}{c}{Student(GPT-3.5) + Learning Scenario} \\
\hline
w/ knowledge & 2.38 & 29.17 & 36.84 & 32.56 & 56.00 & 16.69 & 13.64 \\
same but know 75$\%$ & 2.81 & 29.17 & 31.11 & 30.11 & 52.53 & 13.91 & 11.12 \\
same but know 50$\%$ & 2.38 & 27.08 & 34.21 & 30.23 & 47.22 & 15.91 & 11.95 \\
same but know 25$\%$ & 2.12 & 25.00 & 35.29 & 29.27 & 46.62 & 11.26 & 8.45 \\
\hline
w/ suggestion & 2.06 & 25.00 & 36.36 & 29.63 & 62.39 & 18.43 & 11.49 \\
same but know 75$\%$ & 2.19 & 25.00 & 34.29 & 28.92 & 59.86 & 14.37 & 11.33 \\
same but know 50$\%$ & 1.88 & 20.83 & 33.33 & 25.64 & 48.96 & 9.46 & 7.90 \\
same but know 25$\%$ & 1.69 & 20.83 & 37.04 & 26.67 & 57.29 & 9.20 & 6.25 \\
\hline
w/ both & 3.19 & 33.33 & 31.37 & 32.32 & 49.10 & 13.61 & 10.60 \\
same but know 75$\%$ & 3.62 & 31.25 & 25.86 & 28.30 & 55.80 & 14.58 & 10.81 \\
same but know 50$\%$ & 3.31 & 33.33 & 30.19 & 31.68 & 48.56 & 10.03 & 7.51 \\
same but know 25$\%$ & 3.25 & 31.25 & 28.85 & 30.00 & 50.73 & 7.87 & 5.55 \\
\hline
\multicolumn{8}{c}{Upper bound} \\
\hline
2 Agents & 2.56 & 37.50 & 43.90 & 40.45 & 52.94 & 29.86 & 17.88 \\
\hline
\end{tabular}
\vspace{-2.0em}
\end{table}

\begin{table}
\scriptsize
\caption{The HDE results in the Neurology cases of the department of \textbf{Internal Medicine} with the support of multi-modality.}
\label{multimodal_performance}
\centering
\begin{tabular}{l|ccccc|c}
\hline
 & Symptom & Medical & Diagnostic & Diagnostic & Treatment & Avg\\
 & & Examination & Results & Rationales & Plan & (std)\\
\hline
 & \multicolumn{5}{c}{Student} \\
\hline
GPT-3.5 & 2.562 (0.250) & 1.438 (0.250) & 1.688 (0.312) & 1.625 (0.312) & 1.625 (0.312) & 1.788\\
GPT-4o-mini & 2.688(0.250) & 2.000(0.438) & 2.188(0.500) & 1.875(0.312) & 1.812(0.375) & 2.112 \\
Claude3.5-Sonnet & \textbf{2.875} (0.188) & 2.375 (0.375) & \textbf{2.375} (0.500) & \textbf{2.188} (0.438) & \textbf{2.125} (0.438) & 2.388\\
\hline
& \multicolumn{5}{c}{Student + Multi-modality} \\
\hline
GPT-3.5 & 2.562(0.312) & 1.500(0.250) & 1.750(0.312) & 1.562(0.375) & 1.438(0.250) & 1.762\\
GPT-4o-mini & 2.562(0.250) & 1.938 (0.438) & 2.312(0.500) & 2.000(0.375) & 1.812(0.375) & 2.125\\
Claude3.5-Sonnet & 2.875(0.250) & 2.375(0.438) & 2.312(0.500) & 2.375(0.375) & 2.250(0.438) & 2.437\\
\hline
 & \multicolumn{5}{c}{Upper bound} \\
\hline
2 Agents\cite{fan2024ai} & 2.688(0.250) & 2.125(0.250) & 2.312(0.500) & 1.938(0.375) & 1.750(0.250) & 2.163\\
\hline
\end{tabular}
\vspace{-1.0em}
\end{table}

\begin{table}
\caption{The ICD-10 related results on Neurology cases in the department of \textbf{Internal Medicine Neurology} with multi-modality support}
\label{multimodal_f1}
\centering
\begin{tabular}{l|c|ccc|ccc}
\hline
 & \# & R$\uparrow$ & P$\uparrow$ & F1$\uparrow$ & Coarse($\%$) $\uparrow$ & Medium($\%$) $\uparrow$ & Fine($\%$) $\uparrow$\\
\hline
\multicolumn{8}{c}{Student} \\
\hline
GPT-3.5 & 1.44 & 10.42 & 21.74 & 14.08  &43.86 & 1.39 & 0.00  \\
GPT-4o-mini & 1.56 & 20.83 & 40.00 & 27.40 & 52.95 & 29.60 & 15.02 \\
Claude3.5-Sonnet & 1.62 & 22.92 & \textbf{42.31} & \textbf{29.73}  & 62.32 & \textbf{34.98} & \textbf{11.20} \\
\hline
\multicolumn{8}{c}{Student + Multi-modal} \\
\hline
GPT-3.5 & 1.25 & 16.67 & 40.00 & 23.53 & 55.04 & 19.97 & 8.51 \\
GPT-4o-mini & 1.56 & 22.92 & 44.00 & 30.14 & 46.74 & 27.12 & 19.31 \\
Claude3.5-Sonnet & 1.50 & 29.17 & 58.33 & 38.89 & 59.90 & 31.25 & 19.62 \\

\hline
\multicolumn{8}{c}{Upper bound} \\
\hline
2 Agents & 2.56 & 37.50 & 43.90 & 40.45 & 52.94 & 29.86 & 17.88 \\
\hline
\end{tabular}
\vspace{-2.0em}
\end{table}

\section{Case show}
\begin{figure*}[!th]
    \centering
    \includegraphics[width=1.0\textwidth]{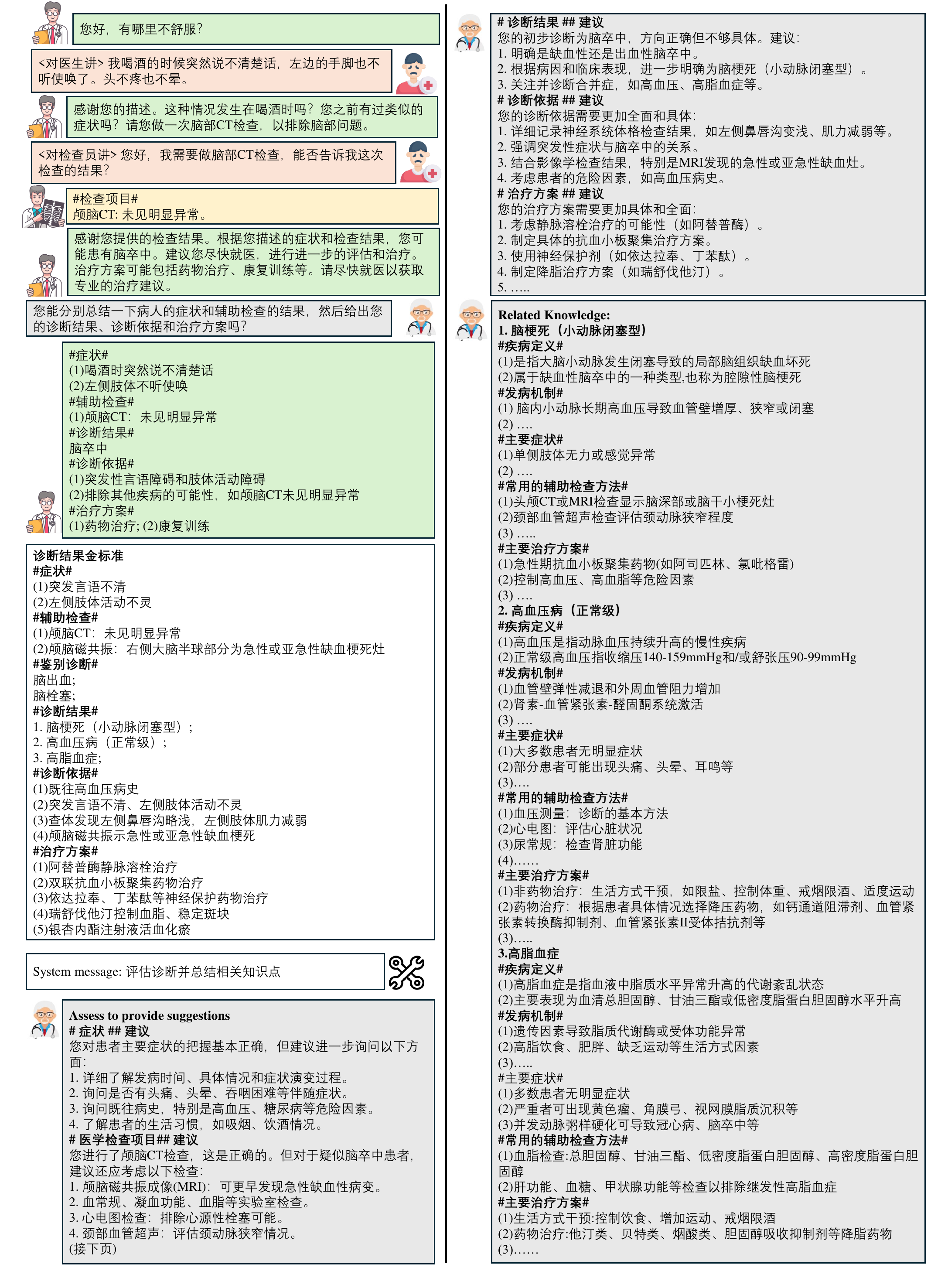}
    \caption{A Chinese example of the learning scenario to acquire the expert's assessment (suggestion) and case-specific knowledge about the involved diseases. The English version is shown in Fig.\ref{fig_learning_case_english}.}
    \label{fig_learning_case_chinese}
\end{figure*}

\begin{figure*}[!th]
    \centering
    \includegraphics[width=1.0\textwidth]{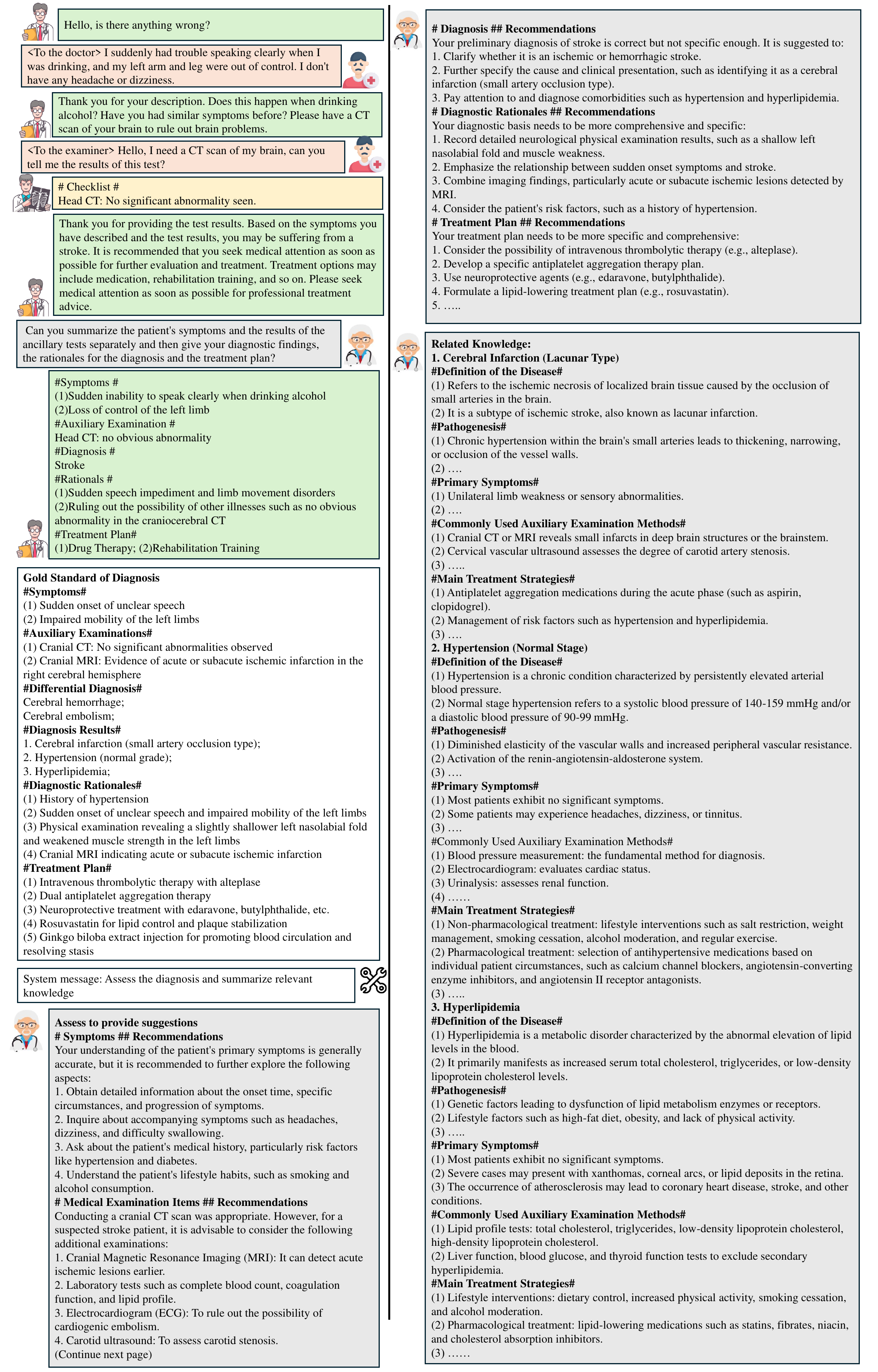}
    \caption{An example of the learning scenario to acquire the expert's assessment (suggestion) and case-specific knowledge about the involved diseases.}
    \label{fig_learning_case_english}
\end{figure*}

\begin{figure*}[!th]
    \centering
    \includegraphics[width=1.0\textwidth]{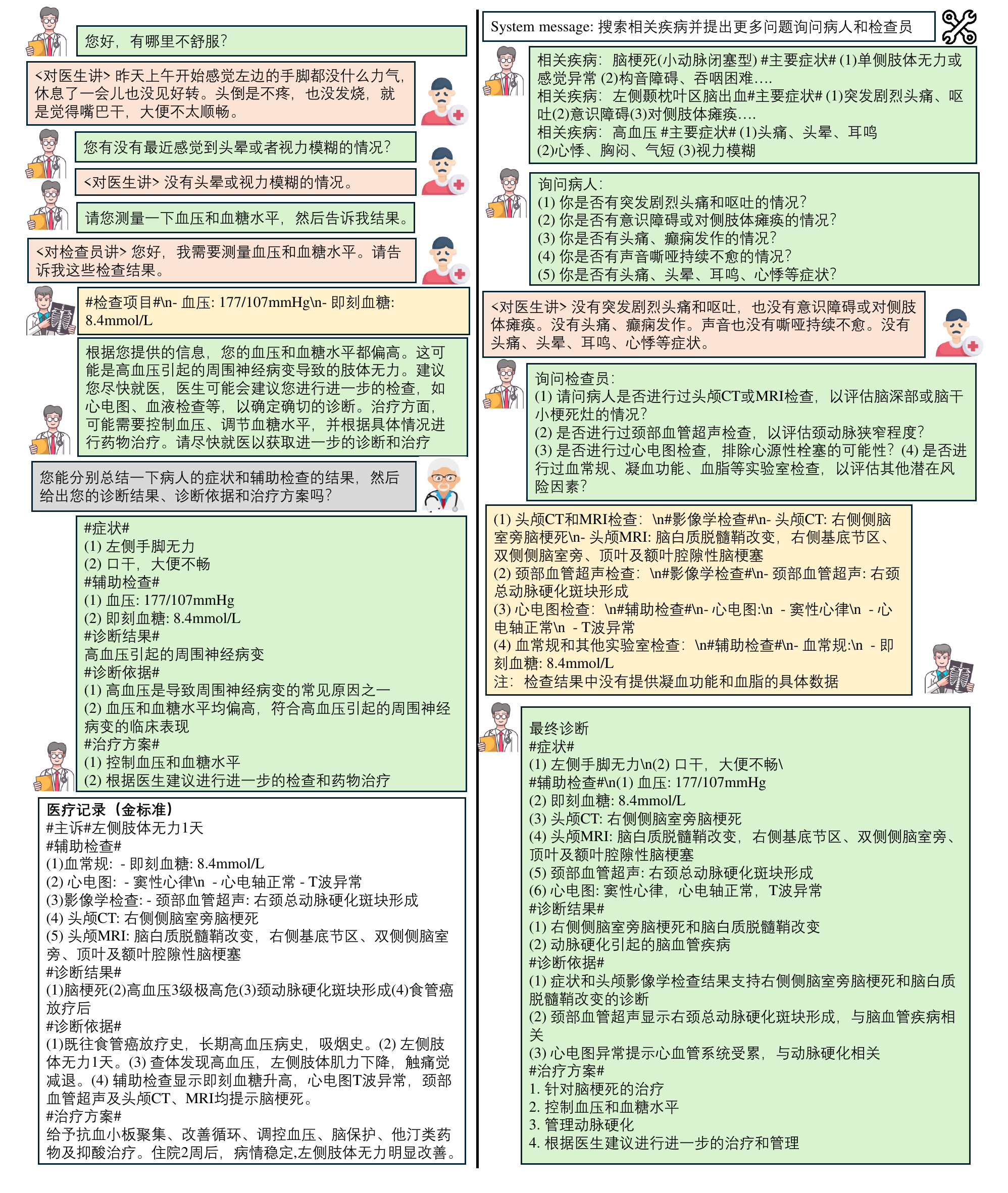}
    \caption{A Chinese example of the practicing scenario where the student retrieves knowledge from the memory to achieve a better diagnosis than the initial one. The English version is shown in Fig.\ref{fig_knowledge_case_english}.}
    \label{fig_knowledge_case_chinese}
\end{figure*}

\begin{figure*}[!th]
    \centering
    \includegraphics[width=1.0\textwidth]{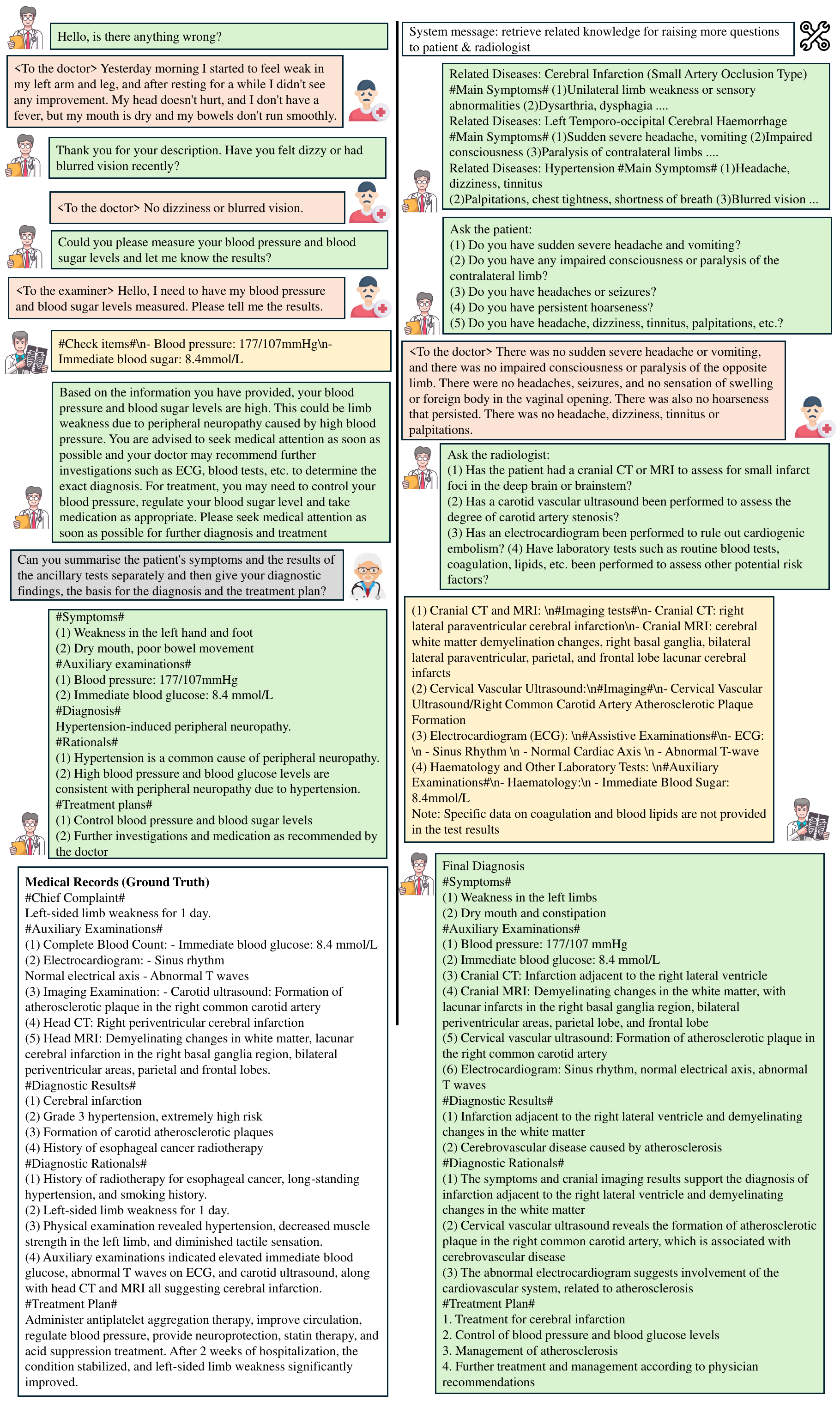}
    \caption{An example of the practicing scenario where the student retrieves knowledge from the memory to achieve a better diagnosis than the initial one.}
    \label{fig_knowledge_case_english}
\end{figure*}

\begin{figure*}[!th]
    \centering
    \includegraphics[width=1.0\textwidth]{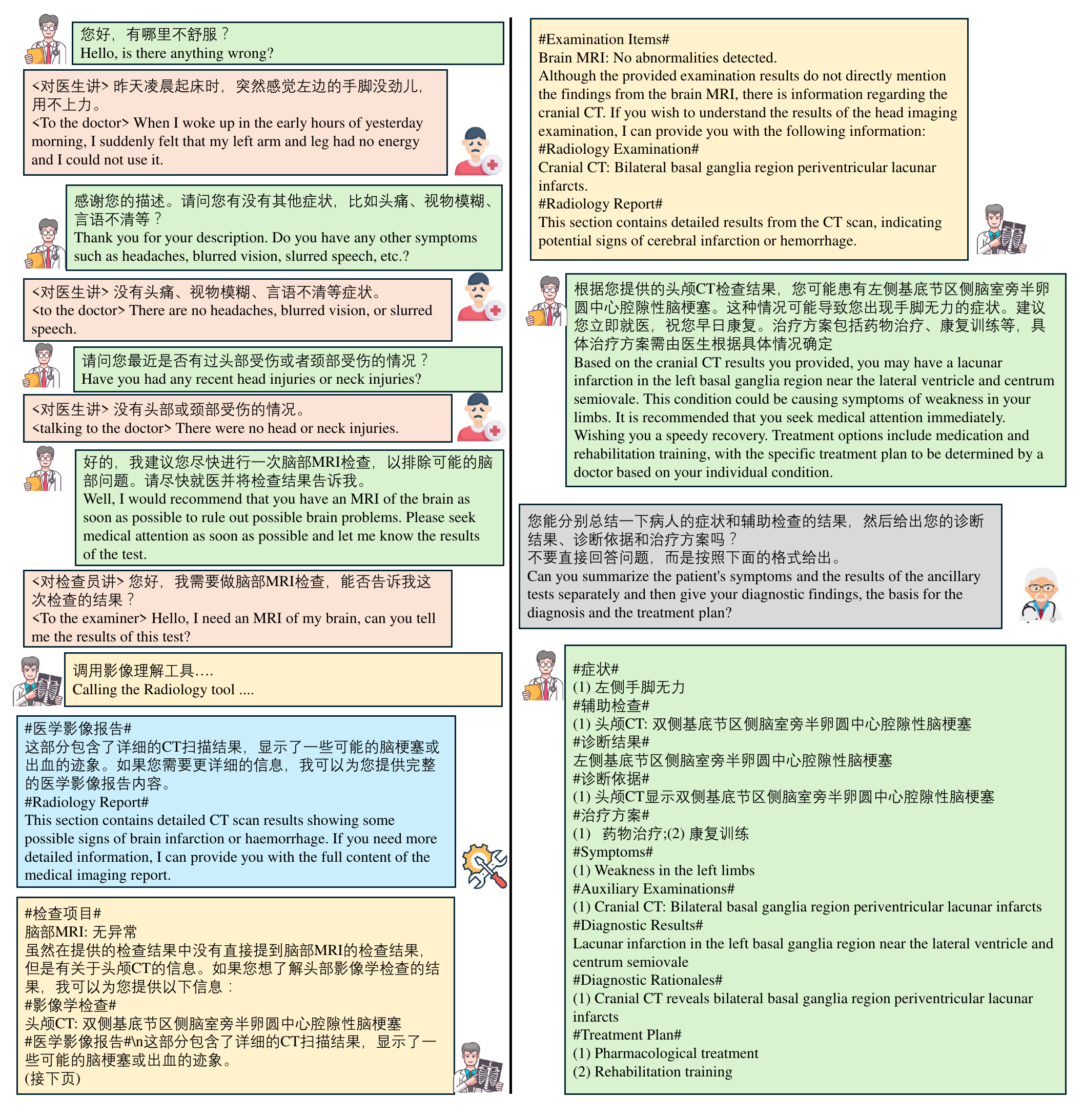}
    \caption{An example of the multi-modal case in practicing scenario where the radiologist calls the radiology tool to interpret the radiological image from the patient to report, where this image is shown in Fig.\ref{fig_multimodal_case_image}.}
    \label{fig_multimodal_case}
\end{figure*}

\begin{figure*}[!th]
    \centering
    \includegraphics[width=1.0\textwidth]{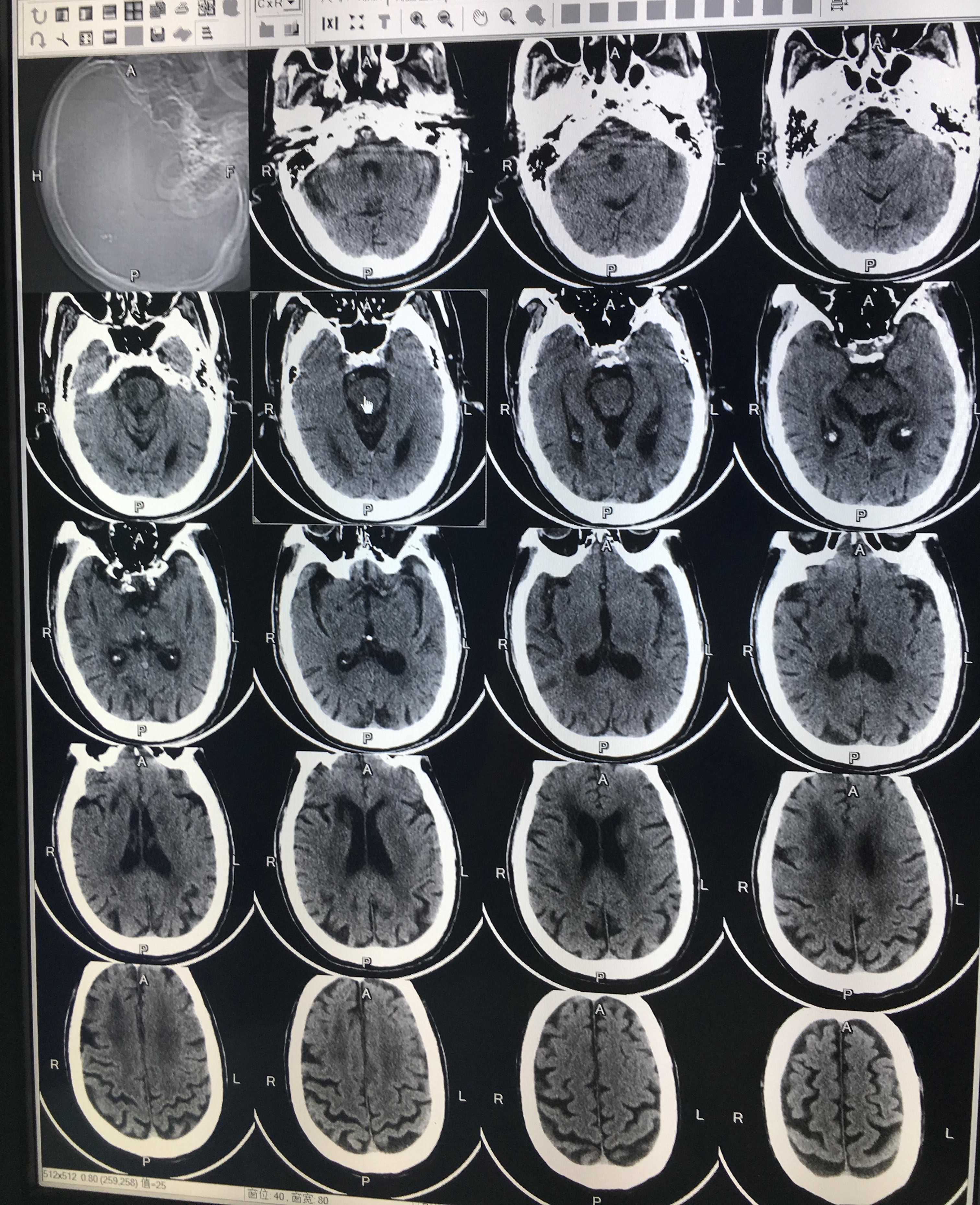}
    \caption{The radiological image of the patient in the multi-modal case}
    \label{fig_multimodal_case_image}
\end{figure*}

This section lists some cases to show the whole interactive diagnosis process, where Fig.\ref{fig_learning_case_chinese} and Fig.\ref{fig_learning_case_english} demonstrate one example in the learning scenario to acquire the suggestions and case-specific knowledge through the assessment of the medical expert. Fig.\ref{fig_knowledge_case_chinese} and Fig.\ref{fig_knowledge_case_english} show how the student utilizes the saved knowledge in the memory to help the diagnosis in the practicing scenario. Fig.\ref{fig_multimodal_case} demonstrates the interactive diagnosis that involves the multi-modality (the image is shown in Fig.\ref{fig_multimodal_case_image})


\begin{table*}[!b]
\begin{tcolorbox}[colback=blue!5!white,colframe=black,width=1.0\textwidth,title={Prompt for the patient role}]
\small
\textbf{System Message}
\vspace{3pt} \\
\begin{CJK}{UTF8}{gbsn}
你是一个病人。这是你的基本资料。\\
\{个性化信息\} \\
\{病历中的基本信息\} 
\end{CJK}
\vspace{5pt} \\
\begin{CJK}{UTF8}{gbsn}
下面会有医生来对你的身体状况进行诊断，\textbf{你需要严格遵循以下指示}：\\
(1) 按照病历和基本资料的设定进行对话。\\
(2) 在每次对话时，你都要明确对话的对象是<医生>还是<检查员>。当你对医生说话时，你要在句子开头说<对医生讲>；如果对象是<检查员>，你要在句子开头说<对检查员讲>。\\
(3) 首先按照主诉进行回复。\\
(4) 当<医生>询问你的现病史、既往史、个人史时，要按照相关内容进行回复。\\
(5) 当<医生>要求或建议你去做检查时，\textbf{不要回答医生，}要立即主动询问<检查员>对应的项目和结果，例如：<对检查员讲> 您好，我需要做xxx检查，能否告诉我这些检查结果？\\
(6) 回答要口语化，尽可能短，提供最主要的信息即可。\\
(7) 从<检查员>那里收到信息之后，将内容主动复述给<医生>。\\
(8) 当医生给出诊断结果、对应的诊断依据和治疗方案后，在对话的末尾加上特殊字符<结束>。
\end{CJK}

\end{tcolorbox}

\begin{tcolorbox}[colback=blue!5!white,colframe=black,width=1.0\textwidth,title={Prompt for the patient role}]
\small
\textbf{System Message}
\vspace{3pt} \\
You are a patient. Here is your basic information.\\
\{Personality\}\\
\{Basic Information in Medical Record\}
\vspace{5pt} \\
A doctor will come to diagnose your physical condition. \textbf{You must strictly follow the following instructions}: \\
(1) Engage in dialogue according to the settings of personality and the basic information in medical record. \\
(2) In each conversation, you must clarify whether you are speaking to a <doctor> or an <examiner>. When you speak to the doctor, you should  start your sentences with <To the doctor>; if the addressee is an <examiner>, you should start with <To the examiner>. \\
(3) First, respond according to the chief complaint. \\
(4) When the <doctor> asks about your present illness history, past medical history, and personal history, reply according to the relevant content. \\
(5) When the <doctor> requests or suggests that you undergo tests, \textbf{Do not answer the doctor,} immediately ask the <examiner> about the corresponding items and results, for example: <To the examiner> Hello, I need to have xxx examination, can you tell me the results of these tests? \\
(6) The responses should be conversational, as short as possible, providing only the most important information. \\
(7) After receiving information from the <examiner>, actively repeat the content to the <doctor>. \\
(8) When the doctor provides the diagnostic result, the corresponding rationale for the diagnosis, and the treatment plan, end the dialogue with the special token <end>. 
\end{tcolorbox}
\caption{The original Chinese and English translated prompts for the patient role, adapted from \cite{fan2024ai} with suitable modification (bold text).}
\label{prompt_patient}
\end{table*}

\begin{table*}[!b]
\begin{tcolorbox}[colback=blue!5!white,colframe=black,width=1.0\textwidth,title={Prompt for medical student in interactive clinical diagnosis (also for initial diagnosis)}]
\small
\textbf{System Message}
\vspace{3pt} \\
\begin{CJK}{UTF8}{gbsn}
你是一个专业且耐心的医生，下面会有患者向你咨询病情。你需要：\\
(1) 在信息不充分的情况下，不要过早作出诊断。\\
(2) 多次、主动地向患者提问来获取充足的信息。\\
(3) \textbf{主动要求患者进行必要的检查}，并等待患者反馈。\\
(4) 诊断结果需要准确到具体疾病。\\
(5) 最后根据患者的身体状况和检查结果，给出诊断结果、对应的诊断依据和治疗方案。\\
\end{CJK}
\end{tcolorbox}

\begin{tcolorbox}[colback=blue!5!white,colframe=black,width=1.0\textwidth,title={Prompt for medical student in interactive clinical diagnosis (also for initial diagnosis)}]
\small
\textbf{System Message}
\vspace{3pt} \\
You are a professional and patient doctor, and you will be consulted by patients. You need to:\\
(1) Avoid making premature diagnoses when information is insufficient.\\
(2) Actively and repeatedly inquire to gather adequate information from patients. \\
(3) \textbf{Proactively request the patient to undergo necessary examinations} and await their feedback. \\
(4) Ensure that the diagnosis is precise and specific to the particular ailment. \\
(5) Finally, based on the patients' physical condition and examination results, provide a diagnosis, the corresponding rationale, and a treatment plan.
\end{tcolorbox}
\caption{The original Chinese and translated English prompts for the medical student in interactive and initial clinical diagnosis adapted from \cite{fan2024ai} with suitable modification (bold text).}
\label{prompt_student1}
\end{table*}

\begin{table*}[!b]
\begin{tcolorbox}[colback=blue!5!white,colframe=black,width=1.0\textwidth,title={Prompt for medical student to start further inquiry conversion}]
\small
\textbf{System Message}
\vspace{3pt} \\
\begin{CJK}{UTF8}{gbsn}
你是一个专业且耐心的医生。你正在为患者做诊断，目前患者的症状和辅助检查如下：\\
\{初步诊断结果\} \\
根据知识库检索，其可能患有这些疾病: {检索到的相关疾病}。\\
为了确定该病人的最终诊断，你继续和患者对话。同时，你需要：\\
(1) 在信息不充分的情况下，不要过早作出诊断。\\
(2) 多次、主动地向患者提问来获取充足的信息。 \\
(3) 每次只提一个问题，尽量简短。\\
(4) 主动要求患者进行必要的检查，并等待患者反馈。\\
(5) 最后根据患者的身体状况和检查结果，给出诊断结果、对应的诊断依据和治疗方案。\\
(6) 诊断结果需要准确到具体疾病，治疗方案中不要包含检查。\\
\end{CJK}
\end{tcolorbox}

\begin{tcolorbox}[colback=blue!5!white,colframe=black,width=1.0\textwidth,title={Prompt for medical student to start further inquiry conversion}]
\small
\textbf{System Message}
\vspace{3pt} \\
You are a professional and patient doctor. You are currently diagnosing a patient, whose symptoms and auxiliary examinations are as follows:\\
\{Initial diagnosis results\} \\
According to the knowledge base retrieval, the patient may be afflicted with the following conditions: {retrieved relevant diseases} \\
To ascertain the patient's definitive diagnosis, you continue to engage in dialogue with the individual. Concurrently, you are required to:\\
(1) In situations where information is lacking, do not hastily arrive at a diagnosis.\\
(2) Proactively and repeatedly engage the patient with inquiries to obtain comprehensive information.\\
(3) Pose one question at a time, keeping it as concise as possible.\\
(4) Proactively request the patient to undergo the necessary examinations and await their feedback.\\
(5) Ultimately, based on the patient's physical condition and examination results, provide a diagnosis, the corresponding rationale for the diagnosis, and a treatment plan. \\
(6) The diagnosis must be precise, identifying the specific illness, while the treatment plan should exclude any mention of further examinations. 
\end{tcolorbox}
\caption{The original Chinese and translated English prompts for the medical student in to start the conversion for the further inquiry stage.}
\label{prompt_student2}
\end{table*}

\begin{table*}[!b]
\begin{tcolorbox}[colback=blue!5!white,colframe=black,width=1.0\textwidth,title={Prompt for medical student to raise more patient-related questions based on the retrieved knowledge}]
\small
\textbf{System Message}
\vspace{3pt} \\
\begin{CJK}{UTF8}{gbsn}
你是一个专业的医生。\\
你正在为患者做诊断，其基本资料为：\{一般资料\} \\ 
患者的症状总结如下：\\
\# 症状\# {初步诊断中的症状信息} \\
根据当前的症状，你将会收到一份知识库检索结果，其中包含与上述症状相关的疾病。 \\
(1) 每种疾病的介绍包含其\#主要症状\#，例如，\#\#相关疾病:高血压\#\#:\#主要症状\# xx \\
(2) 你需要对比分析不同疾病\#主要症状\#之间的相同点和不同点  \\
(3) 根据不同疾病\#主要症状\#之间的不同点，请以\#询问病人\#开头，简明扼要地询问病人更多信息以鉴别区分不同疾病  \\
请你直接按照下面的格式来进行输出，不要回答其他内容： \\
\#询问病人\#(1) xx (2) xx ... \\
\vspace{5pt} \\
\textbf{User}
\vspace{3pt} \\
\{\#\#相关疾病:{xxx}\#\# \#主要症状\#{xxx}\}
\end{CJK}
\end{tcolorbox}

\begin{tcolorbox}[colback=blue!5!white,colframe=black,width=1.0\textwidth,title={Prompt for medical student to raise more patient-related questions based on the retrieved knowledge}]
\small
\textbf{System Message}
\vspace{3pt} \\
You are a professional doctor.\\
You are conducting a diagnosis for the patient, whose basic information is as follows: {General Information} \\
The patient's symptoms are summarized as follows:\\
\# Symptoms\# {the symptoms in the initial diagnosis} \\
Based on the current symptoms, you will receive a knowledge base retrieval result, which includes diseases associated with the aforementioned symptoms. \\
(1) Each disease description includes its \#primary symptoms\#, for example, \#\#Related Disease: Hypertension\#\#:\#Primary Symptoms\# xx. \\
(2) You are required to conduct a comparative analysis of the similarities and differences among the \#primary symptoms\# of various diseases. \\
(3) Beginning with \#Inquire Patient\#, succinctly ask the patient for additional information to discern and differentiate between the various diseases based on the differences in their \#primary symptoms\#. \\
Please follow the format below for output, without providing any additional content: \\
\# Inquire Patient\#(1) xx (2) xx ... \\
\vspace{5pt} \\
\textbf{User}
\vspace{3pt} \\
\{\#\#Related Disease:{xxx}\#\# \#Primary Symptoms\#{xxx}\}
\end{tcolorbox}
\caption{Original Chinese and English prompts for medical students to inquire about patients based on retrieved knowledge.}
\label{prompt_student3}
\end{table*}

\begin{table*}[!b]
\begin{tcolorbox}[colback=blue!5!white,colframe=black,width=1.0\textwidth,title={Prompt for medical student to raise more radilogist-related questions based on the retrieved knowledge}]
\small
\textbf{System Message}
\vspace{3pt} \\
\begin{CJK}{UTF8}{gbsn}
你是一个专业的医生。\\
你正在为患者做诊断，其基本资料为：\{一般资料\} \\ 
患者的症状辅助检查总结如下：\\
\# 症状\# {初步诊断中的症状信息} \\
\# 辅助检查\# {初步诊断中的辅助检查信息} \\
根据当前的症状，你将会收到一份知识库检索结果，其中包含与上述症状相关的疾病。 \\
(1) 每种疾病的介绍包含其\#常用辅助检查方法\#，例如影像学检查，头颅CT等 \\
(2) 你需要对比分析不同疾病所需要的辅助检查方法的相同点和不同点 \\
(3) 根据这些辅助检查方法，请以\#询问检查员\#开头，简明扼要地询问更多当前病人的辅助检查信息以鉴别区分不同疾病  \\
请你直接按照下面的格式来进行输出，不要回答其他内容： \\
\#询问病人\#(1) xx (2) xx ... \\
\vspace{5pt} \\
\textbf{User}
\vspace{3pt} \\
\{\#\#相关疾病:{xxx}\#\# \#常用辅助检查方法\#{xxx}\}
\end{CJK}
\end{tcolorbox}

\begin{tcolorbox}[colback=blue!5!white,colframe=black,width=1.0\textwidth,title={Prompt for medical student to raise more radiologist-related questions based on the retrieved knowledge}]
\small
\textbf{System Message}
\vspace{3pt} \\
You are a professional doctor.\\
You are conducting a diagnosis for the patient, whose basic information is as follows: {General Information} \\
The patient's symptoms and auxiliary examinations are summarized as follows:\\
\# Symptoms\# {the symptoms in the initial diagnosis} \\
\# Examinations \# {the examinations in the initial diagnosis}
Based on the current symptoms, you will receive a knowledge base retrieval result, which includes diseases associated with the aforementioned symptoms. \\
(1) Each disease description includes its \#commonly used auxiliary examination methods\#, such as imaging studies, cranial CT scans, and so forth.\\
(2) You are required to conduct a comparative analysis of the similarities and differences in the auxiliary examination methods required for various diseases. \\
(3) Beginning with \#Inquire Radiologist\#, succinctly request additional information regarding the current patient's auxiliary examinations to discern and differentiate between the various diseases. \\
Please follow the format below for output, without providing any additional content: \\
\# Inquire Radiologist\#(1) xx (2) xx ... \\
\vspace{5pt} \\
\textbf{User}
\vspace{3pt} \\
\{\#\#Related Disease:{xxx}\#\# \#commonly used auxiliary examination methods\#{xxx}\}
\end{tcolorbox}
\caption{Original Chinese and English prompts for medical students to inquire about radiologist based on retrieved knowledge.}
\label{prompt_student4}
\end{table*}

\begin{table*}[!b]
\begin{tcolorbox}[colback=blue!5!white,colframe=black,width=1.0\textwidth,title={Prompt for medical student to discuss and fuse the diagnosis from knowledge-based and suggestion-based inquiry.}]
\small
\textbf{System Message}
\vspace{3pt} \\
\begin{CJK}{UTF8}{gbsn}
你是一个资深的\#主任医生\#。\\
你正在主持一场医生针对患者病情的会诊，参与的医生有\{医生A和医生B\}。 \\ 
(1) 你需要听取每个医生的诊断报告，其中包含对病人的\#症状\#、\#辅助检查\#、\#诊断结果\#、\#诊断依据\#和\#治疗方案\#。 \\
(2) 你需要汇总每个医生的信息，给出对病人的最终诊断。 \\
(3) 在汇总诊断结果的时候，确保简明扼要。根据需要可适当合并,而不是拼接不同医生的结果  \\
(4) 请你按照下面的格式来进行输出 \\
请你直接按照下面的格式来进行输出，不要回答其他内容： \\
\#症状\# (1) xxx (2) xxx \\
\#辅助检查\# (1) xxx (2) xxx \\
\#诊断结果\# (1) xxx (2) xxx \\
\#诊断依据\# (1) xxx (2) xxx \\
\#治疗方案\# (1) xxx (2) xxx \\
\vspace{5pt} \\
\textbf{User}
\vspace{3pt} \\
\{\#\#医生A\#\# 诊断报告....\}
\end{CJK}
\end{tcolorbox}

\begin{tcolorbox}[colback=blue!5!white,colframe=black,width=1.0\textwidth,title={Prompt for medical student to discuss and fuse the diagnosis from knowledge-based and suggestion-based inquiry}]
\small
\textbf{System Message}
\vspace{3pt} \\
You are a seasoned \#Experienced Doctor\#\\
You are presiding over a consultation among physicians regarding the patient's condition, with participants including \{Doctor A and Doctor B\}.\\
(1) You need to listen to each doctor's diagnostic report, which encompasses the patient's \#symptoms\#, \#examinations\#, \#diagnostic results\#, \#rationales\#, and \#treatment plan\#.\\
(2) You are required to compile the information from each physician and provide a conclusive diagnosis for the patient. \\
(3) When consolidating the diagnostic results, ensure that the summary is concise and succinct. Merge the findings as necessary, rather than merely concatenating the different doctor' results. \\
(4) Please present the information in the following format. \\
Please follow the format below for your output, without providing any additional content: \\
\#Symptom\# (1) xxx (2) xxx \\
\#Examinations\# (1) xxx (2) xxx \\
\#Diagnostic Results\# (1) xxx (2) xxx \\
\#Rationale\# (1) xxx (2) xxx \\
\#Treatment Plan \# (1) xxx (2) xxx \\
\vspace{5pt} \\
\textbf{User}
\vspace{3pt} \\
\{\#\#Doctor A\#\# Diagnostic report \}
\end{tcolorbox}
\caption{Original Chinese and English prompts for medical students to collaboratively discuss and integrate diagnoses through knowledge-based and suggestion-based inquiry.}
\label{prompt_student5}
\end{table*}

\begin{table*}[!b]
\begin{tcolorbox}[colback=blue!5!white,colframe=black,width=1.0\textwidth,title={Prompt for the radiologist to handle examination request.}]
\small
\textbf{System Message}
\vspace{3pt} \\
\begin{CJK}{UTF8}{gbsn}
你是医院的数据库管理员，负责收集、汇总和整理病人的病史和检查数据。\\
这是你收到的病人的检查结果。\\
\#查体\#{Examination section中的查体信息} \\ 
\#辅助检查\#{Examination section中的辅助检查信息} \\
\#医学影像报告\#{调用工具得到的图像的文字描述，如果适用} \\
下面会有病人或者医生来查询，你要忠实地按照收到的检查结果，找到对应的项目，并按照下面的格式来回复。\\
\#检查项目\#- xxx: xxx - xxx: xxx\#xx检查\# - xxx: xxx - xxx: xxx \\
如果无法查询到对应的检查项目则回复：\\
- xxx: 无异常
\vspace{5pt} \\
\textbf{User}
\vspace{3pt} \\
\{您好，我需要做基因组测序，能否告诉我这些检查结果？\}
\textbf{Assistant}[Radiologist]
\vspace{3pt} \\
\{\#检查项目\#- 基因组测序\}
\vspace{5pt} \\
\textbf{User} [患者]
\vspace{3pt} \\
\{您好，我需要做颈部超声检查和血常规检查，请告诉我这些检查结果？\}
\end{CJK}
\end{tcolorbox}

\begin{tcolorbox}[colback=blue!5!white,colframe=black,width=1.0\textwidth,title={Prompt for the radiologist to handle examination request.}]
\small
\textbf{System Message}
\vspace{3pt} \\
As the database administrator of the hospital, you are responsible for collecting, consolidating, and organizing patients' medical histories and examination data.\\
Here are the examination results you have received for the patient.\\
\#Physical Examination\#{the Physical Examination in the Examination section} \\ 
\#Auxiliary Examinations\#{ the Auxiliary Examinations in the Examination section} \\
\#Medical Imaging Report\# \{Textual description of the images obtained from the tool, if applicable\} \\
There will be patients or doctors inquiring below; you are to faithfully refer to the received examination results, locate the corresponding items, and respond in the following format. \\
\#Examination Items\# - xxx: xxx - xxx: xxx \#xx Examination\# - xxx: xxx - xxx: xxx \\
If the corresponding examination item cannot be found, please respond with: \\
xxx: No abnormalities detected.
\vspace{5pt} \\
\textbf{User} \\
\vspace{3pt} \\
\{Hello, I would like to request the results of the genomic sequencing. Could you kindly provide me with this information?\}
\textbf{Assistant}[Radiologist]
\vspace{3pt} \\
\{\#Examination Items\# - Genomic Sequencing\}
\vspace{5pt} \\
\textbf{User} [Patient]
\vspace{3pt} \\
\{Hello, I require the results for the cervical ultrasound and complete blood count examinations. Could you kindly provide me with this information?\}
\end{tcolorbox}
\caption{Original Chinese and English prompts for the radiologist to handle examination requests from patients and medical students.}
\label{prompt_radiologist1}
\end{table*}

\begin{table*}[!b]
\begin{tcolorbox}[colback=blue!5!white,colframe=black,width=1.0\textwidth,title={Prompt for selecting suitable tools to interpret the upload images}]
\small
\textbf{System Message}
\vspace{3pt} \\
\begin{CJK}{UTF8}{gbsn}
你是一个智能医疗助手，负责处理各种医学相关的图像。\\
你的任务是根据用户上传的图像类型，选择合适的工具进行处理。请按照以下步骤操作：\\
(1) 仔细观察用户上传的图像\\
(2) 判断图像类型并根据类型来选择合适的工具。\\
(2.1) 如果输入的是医学影像，如如X光片、CT扫描、MRI等，则选择相关工具进行报告生成\\
(2.2) 如果输入的是医学报告的照片或扫描件，则选择相关工具进行报告解读\\
(3) 如果是其他类型的图像，如自然风光、人物等，则直接输出\#无\#，不需要进行解读\\
请记住，你的角色是协助选择合适的工具，而不是直接生成报告或解读报告。如果用户有任何疑问，请耐心解答。
\vspace{5pt} \\
\textbf{User}
\vspace{3pt} \\
\{图片路径：{xxx}, base64 code of the image\}
\end{CJK}
\end{tcolorbox}

\begin{tcolorbox}[colback=blue!5!white,colframe=black,width=1.0\textwidth,title={Prompt for selecting suitable tools to interpret the upload images.}]
\small
\textbf{System Message}
\vspace{3pt} \\
You are an intelligent medical assistant responsible for processing various medical-related images.\\
Your task is to select the appropriate tool for processing based on the type of image uploaded by the user. Please follow these steps: \\
(1) Carefully observe the image uploaded by the user. \\
(2) Determine the type of image and select the appropriate tool accordingly. \\
(2.1) If the input is a medical image, such as an X-ray, CT scan, MRI, etc., choose the relevant tool to generate a report. \\
(2.2) If the input is a photograph or scanned copy of a medical report, select the relevant tool for report interpretation. \\
(3) If it is another type of image, such as landscapes or portraits, simply output \#None\#, without any interpretation. \\
Please remember, your role is to assist in selecting the appropriate tool, not to directly generate reports or interpret them. If the user has any questions, please respond patiently.
\vspace{5pt} \\
\textbf{User}
\vspace{3pt} \\
\{\#\#The path of image: \{xxx\}, base64 code of the image \}
\end{tcolorbox}
\caption{Original Chinese and English prompts to select the suitable tools to interpret the uploaded images.}
\label{prompt_radiologist2}
\end{table*}

\begin{table*}[!b]
\begin{tcolorbox}[colback=blue!5!white,colframe=black,width=1.0\textwidth,title={Prompt for radiology tool to interpret radiological images as report}]
\small
\textbf{System Message}
\vspace{3pt} \\
\begin{CJK}{UTF8}{gbsn}
你是一位经验丰富的放射科医生，专门解读各种医学影像。我将为你提供一张医学影像图片。\\
你的任务是仔细分析这张图片，并生成一份专业的影像报告。请遵循以下指南：\\
(1) 首先，识别并说明这是什么类型的医学影像（如X光、CT、MRI等） \\
(2) 报告格式应包含：检查类型、技术细节（如图像质量、拍摄位置等）、详细发现和结论。\\
(3) 在详细发现部分，请描述：可见的解剖结构及其正常/异常情况、任何病变或异常的具体位置、大小、形状和特征、密度、信号强度或对比度的异常周围组织的情况\\
(4) 在结论部分：总结主要发现、提供可能的诊断或鉴别诊断。 \\
(5) 如果描述中的某些细节不清楚或缺失，请在报告中注明\\
(6) 报告应该客观、准确，避免推测性的结论。\\
请仔细观察提供的医学影像图片，然后基于你的观察生成一份专业的影像报告.
\vspace{3pt} \\
\textbf{User}
\vspace{3pt} \\
\{图片路径：{xxx}, base64 code of the image\}
\vspace{3pt} \\
\textbf{工具描述}
\vspace{3pt} \\
对输入的医学影像进行解读并生成的专业的影像报告，其包括图像类型、技术细节、具体发现和结论等。只接受医学影像图片，其他类型图像，如人物、自然图像等将不会进行解读
\end{CJK}
\end{tcolorbox}

\begin{tcolorbox}[colback=blue!5!white,colframe=black,width=1.0\textwidth,title={Prompt for radiology tool to interpret radiological images as report.}]
\small
\textbf{System Message}
\vspace{3pt} \\
You are an experienced radiologist specializing in the interpretation of various medical images. I will provide you with a medical imaging picture.\\
Your task is to carefully analyze this image and generate a professional imaging report. Please adhere to the following guidelines: \\
(1) First, identify and specify the type of medical imaging (such as X-ray, CT, MRI, etc.). \\
(2) The report format should include: the type of examination, technical details (such as image quality, acquisition position, etc.), detailed findings, and conclusion. \\
(3) In the detailed findings section, please describe the visible anatomical structures and their normal/abnormal conditions, the specific location, size, shape, and characteristics of any lesions or abnormalities, density, signal intensity, or contrast abnormalities - the condition of surrounding tissues. \\
(4) In the conclusion section: summarize the main findings and provide possible diagnoses or differential diagnoses. \\
(5) If certain details in the description are unclear or missing, please note this in the report. \\
(6) The report should be objective and accurate, avoiding speculative conclusions. \\
Please carefully observe the provided medical imaging picture and generate a professional imaging report based on your observations. \\
\vspace{1pt} \\
\textbf{User}
\vspace{1pt} \\
\{\#\#The path of image: \{xxx\}, base64 code of the image \}
\vspace{1pt} \\
\textbf{Tool's description}
\vspace{1pt} \\
Interpretation of the provided medical images will result in a professional imaging report that includes image type, technical details, specific findings, and conclusions. Only medical images will be accepted; other types of images, such as portraits or nature photos, will not be interpreted.
\end{tcolorbox}
\caption{Original Chinese and English prompts radiology tool to interpret radiological images as the report.}
\label{prompt_tool1}
\end{table*}

\begin{table*}[!b]
\begin{tcolorbox}[colback=blue!5!white,colframe=black,width=1.0\textwidth,title={Prompt for ReportVQA tool to interpret radiological report photos as textual description}]
\small
\textbf{System Message}
\vspace{3pt} \\
\begin{CJK}{UTF8}{gbsn}
你是一位经验丰富的医疗记录分析专家，专门解读各种医学影像报告。\\
现在有一张医学影像报告的照片需要你仔细分析。请按照以下步骤进行解读：\\
(1) 首先描述你看到的报告类型（如X光报告、CT扫描报告、MRI报告等）和涉及的身体部位。如果看不清，可不描述该项内容 \\
(2) 然后，详细解读报告中的各项内容，包括但不限于：检查类型、技术细节（如图像质量、拍摄位置等）、详细发现和结论。\\
(3) 指出报告中任何不清楚、模糊或可能需要进一步澄清的部分。\\
请记住，你的解读应该准确、全面且易于理解。如果报告中有任何不确定或需要进一步解释的内容，请明确指出。\\
现在，请仔细查看提供的医学影像报告照片，并按照上述步骤进行详细解读。
\vspace{5pt} \\
\textbf{User}
\vspace{3pt} \\
\{图片路径：{xxx}, base64 code of the image\}
\vspace{5pt} \\
\textbf{工具描述}
\vspace{3pt} \\
对输入的医学影像的报告照片进行解读，包括报告类型、报告内容和结论等。只接受影像报告的照片，其他类型图像，如CT、自然图像等将不会进行解读
\end{CJK}
\end{tcolorbox}

\begin{tcolorbox}[colback=blue!5!white,colframe=black,width=1.0\textwidth,title={Prompt for ReportVQA tool to interpret radiological report photos as textual description}]
\small
\textbf{System Message}
\vspace{3pt} \\
You are an experienced medical record analysis expert, specializing in the interpretation of various medical imaging reports.\\
Now, there is a photograph of a medical imaging report that requires your careful analysis. Please follow these steps for interpretation:\\
(1) Firstly, describe the type of report you observe (such as X-ray report, CT scan report, MRI report, etc.) and the relevant body parts. If the details are unclear, do not describe that aspect.\\
(2) Next, provide a detailed interpretation of the contents of the report, including but not limited to: the type of examination, technical details (such as image quality, positioning, etc.), detailed findings, and conclusions.\\
(3) Point out any unclear, ambiguous, or potentially needing further clarification sections within the report.\\
Please remember that your interpretation should be accurate, comprehensive, and easy to understand. If there are any uncertainties or elements that require further elaboration, please indicate them clearly.\\
Now, please carefully examine the provided medical imaging report photograph and proceed with a detailed interpretation according to the aforementioned steps.
\vspace{5pt} \\
\textbf{User}
\vspace{3pt} \\
\{\#\#The path of image: \{xxx\}, base64 code of the image \}
\vspace{5pt} \\
\textbf{Tool's description}
\vspace{3pt} \\
Interpret the report photos of the input medical images, including the type of report, report content, and conclusions. Only photos of imaging reports will be accepted; other types of images, like CT scans or natural images, will not be interpreted.
\end{tcolorbox}
\caption{Original Chinese and English prompts for the ReportVQA tool to interpret radiological report photos as textual descriptions.}
\label{prompt_tool2}
\end{table*}

\begin{table*}[!b]
\begin{tcolorbox}[colback=blue!5!white,colframe=black,width=1.0\textwidth,title={Prompt for the medical expert to ask the student to summarize the diagnosis}]
\small
\textbf{System Message}
\vspace{3pt} \\
\begin{CJK}{UTF8}{gbsn}
您能分别总结一下病人的症状和辅助检查的结果，然后给出您的诊断结果、诊断依据和治疗方案吗？\\
不要直接回答问题，而是按照下面的格式给出。\\
\#症状\# (1)xx (2)xx \\
\#辅助检查\# (1)xx (2)xx \\
\#诊断结果\# xx \\
\#诊断依据\# (1)xx (2)xx \\
\#治疗方案\# (1)xx (2)xx \\
\end{CJK}
\end{tcolorbox}

\begin{tcolorbox}[colback=blue!5!white,colframe=black,width=1.0\textwidth,title={Prompt for the medical expert to ask the student to summarize the diagnosis}]
\small
\textbf{System Message}
\vspace{3pt} \\
Please summarize the patient's symptoms and the results of auxiliary examinations separately, and then provide your diagnostic conclusion, the basis for your diagnosis, and the treatment plan.\\
Do not answer the questions directly; instead, present your response in the following format.\\
\#Symptom\# (1) xxx (2) xxx \\
\#Examinations\# (1) xxx (2) xxx \\
\#Diagnostic Results\# (1) xxx (2) xxx \\
\#Rationale\# (1) xxx (2) xxx \\
\#Treatment Plan \# (1) xxx (2) xxx \\
\end{tcolorbox}
\caption{Original Chinese and English prompts for the medical expert to ask the student to summarize the diagnosis.}
\label{prompt_expert1}
\end{table*}

\begin{table*}[!b]
\begin{tcolorbox}[colback=blue!5!white,colframe=black,width=1.0\textwidth,title={Prompt for the medical expert to assess the diagnosis of the student}]
\small
\textbf{System Message}
\vspace{3pt} \\
\begin{CJK}{UTF8}{gbsn}
作为一名经验丰富的的医学专家。\\
请你根据专家诊疗结果中的现病史、辅助检查、诊断结果、诊断依据和治疗方案，对实习医生的诊疗过程进行指导和点评。\\
通过下面的方式来呈现结果\\
\#症状\#\# 建议<根据专家记录的病人病史，为医学上后续与病人对话中有关症状询问提出建议>\\
\#医学检查项目\#\# 建议<基于专家所做的医学检查项目，为实习医生后续与病人对话中有关医学检查项目提出建议>\\
\#诊断结果\#\# 建议<基于专家做出的诊断结果，结合你的医学常识，为实习医生后续诊断过程提出建议>\\
\#诊断依据\#\# 建议<对比专家的诊断依据，为医学生后续分析诊断依据提出建议>\\
\#治疗方案\#\# 建议<对比专家的治疗方案，为医学生后续给病人安排治疗方案时提出建议>\\
(1) 请以专业、耐心且富有教育意义的口吻给出您的反馈，帮助实习医生提高临床诊断能力。\\
(2) 请侧重医学答案的事实内容，不需关注风格、语法、标点和无关医学的内容。\\
(3) 请不要直接回答问题，而是给出建议。
\vspace{5pt} \\
\textbf{User}
\vspace{3pt} \\
\{\#诊断结果label\# \{xxx\}  \#医生诊断结果\# \{xxx\}\}
\end{CJK}
\end{tcolorbox}

\begin{tcolorbox}[colback=blue!5!white,colframe=black,width=1.0\textwidth,title={Prompt for the medical expert to assess the diagnosis of the student}]
\small
\textbf{System Message}
\vspace{3pt} \\
As an experienced medical expert, please guide and provide feedback on the clinical process of the medical student based on the expert evaluation results, including the current medical history, auxiliary examinations, diagnostic results, diagnostic basis, and treatment plan. Present the results as follows:\\
\#Symptoms\#\# Suggestions<Provide suggestions for the student regarding inquiries about symptoms based on the patient's medical history recorded by the expert>\\
\#Medical Examination Items\#\# Suggestions<Offer suggestions for the student regarding medical examination items based on what the expert conducted>
\#Diagnostic Results\#\# Suggestions<Provide suggestions for the student regarding the diagnostic process, utilizing the expert’s diagnostic results and your medical knowledge> \\
\#Rational\#\# Suggestions<Suggest approaches for the student to analyze the diagnostic basis by comparing it with the expert’s diagnostic basis>\\
\#Treatment Plan\#\# Suggestions<Advise the student on arranging treatment plans for the patient by comparing it with the expert’s treatment plan>\\
(1) Please provide your feedback in a professional, patient, and educational manner to help the student improve their clinical diagnostic abilities.\\
(2) Focus on the factual content of medical answers without concern for style, grammar, punctuation, or unrelated medical topics.\\
(3) Do not directly answer the questions, but instead offer suggestions.\\
\vspace{5pt} \\
\textbf{User}
\vspace{3pt} \\
\{\#Ground truth of the diagnosis \# \{xxx\}  \#Diagnosis of expert\# \{xxx\}\}
\end{tcolorbox}
\caption{Original Chinese and English prompts for the medical expert to assess the diagnosis of the student}
\label{prompt_expert2}
\end{table*}

\begin{table*}[!b]
\begin{tcolorbox}[colback=blue!5!white,colframe=black,width=1.0\textwidth,title={Prompt for the medical expert to summarize the case-specific knowledge}]
\small
\textbf{System Message}
\vspace{3pt} \\
\begin{CJK}{UTF8}{gbsn}
你是资深的医学专家 \\
请根据所提供的疾病名称，如高血压，按照下面的格式提供该疾病的相关信息：\\
\#疾病定义\# (1)xx (2)xx \\
\#发病机制\# (1)xx (2)xx \\
\#主要症状\# (1)xx (2)xx \\
\#常用的辅助检查方法\# (1)xx (2)xx \\
\#主要治疗方案\# (1)xx (2)xx \\
(1) 请侧重所提供病例的事实内容，不需关注风格、语法、标点和无关医学的内容。\\
(2) 请你充分利用医学知识，分析并总结每个点的主要内容。\\
(3) 注意诊断结果、诊断依据和治疗方案三者之间的承接关系。\\
(4) 请不要直接回答问题，而是将疾病相关信息总结为上面的格式。
\vspace{5pt} \\
\textbf{User}
\vspace{3pt} \\
\{\#疾病名称：\{xxx\}\}
\end{CJK}
\end{tcolorbox}

\begin{tcolorbox}[colback=blue!5!white,colframe=black,width=1.0\textwidth,title={Prompt for the medical expert to summarize the case-specific knowledge}]
\small
\textbf{System Message}
\vspace{3pt} \\
You are an experienced medical expert. Please provide relevant information about the specified disease, like hypertension, following the format below:\\
\#Disease Definition\# (1) xx (2) xx \\
\#Pathogenesis\# (1) xx (2) xx\\
\#Main Symptoms\# (1) xx (2) xx\\
\#Common Auxiliary Examination Methods\# (1) xx (2) xx\\
\#Main Treatment Plans\# (1) xx (2) xx\\
(1) Focus on the factual content of the provided case without worrying about style, grammar, punctuation, or unrelated medical content.\\
(2) Utilize your medical knowledge to analyze and summarize the key points for each section.\\
(3) Pay attention to the connections between the diagnosis results, diagnostic basis, and treatment plans.\\
(4) Please do not directly answer questions, but summarize the disease-related information in the format above.\\
\vspace{5pt} \\
\textbf{User}
\vspace{3pt} \\
\{\#Disease's name:\{xxx\}\}
\end{tcolorbox}
\caption{Original Chinese and English prompts for the medical expert to summarize the case-specific knowledge}
\label{prompt_expert3}
\end{table*}

\end{document}